\documentclass[10pt,twocolumn,letterpaper]{article}

\usepackage{cvpr}              % To produce the CAMERA-READY version
\usepackage{adjustbox}
\usepackage{amsmath}
\usepackage[accsupp]{axessibility}

\usepackage{physics}
\usepackage{multirow}

\newcommand\T{\rule{0pt}{2.5ex}}       % Top strut
\newcommand\B{\rule[-1.2ex]{0pt}{0pt}} % Bottom strut

\newcommand\blfootnote[1]{%
  \begingroup
  \renewcommand\thefootnote{}\footnote{#1}%
  \addtocounter{footnote}{-1}%
  \endgroup
}

\usepackage{makecell}

\usepackage[dvipsnames]{xcolor}

\definecolor{cvprblue}{rgb}{0.21,0.49,0.74}
\usepackage[pagebackref,breaklinks,colorlinks,citecolor=cvprblue]{hyperref}

\def\paperID{7204} % *** Enter the Paper ID here
\def\confName{CVPR}
\def\confYear{2024}

\title{Model Inversion Robustness: Can Transfer Learning Help?}

\author{%
  Sy-Tuyen Ho$^{1}$\qquad
  Koh Jun Hao$^{1}$\qquad \\
  Keshigeyan Chandrasegaran$^{2}\dag$\qquad
  Ngoc-Bao Nguyen$^{1}$ \qquad
  Ngai-Man Cheung$^{1}$ \\
  $^1$Singapore University of Technology and Design (SUTD)\qquad
  $^2$Stanford University \and
  \texttt{hosy\_tuyen@sutd.edu.sg}\quad
  \texttt{ngaiman\_cheung@sutd.edu.sg}
}

\begin{document}
\maketitle

\blfootnote{$\dag$ Work done while at SUTD.}

\begin{abstract}

Model Inversion (MI) attacks aim to reconstruct private training data by abusing access to machine learning models. Contemporary MI attacks have achieved impressive attack performance, posing serious threats to privacy. Meanwhile, all existing MI defense methods rely on regularization that is in direct conflict with the training objective, resulting in noticeable degradation in model utility. \textbf{In this work,} we take a different perspective, and  propose a novel and simple \underline{T}ransfer \underline{L}earning-based \underline{D}efense against \underline{M}odel \underline{I}nversion (TL-DMI) to render MI-robust models. Particularly, by leveraging TL, we limit the number of  layers encoding sensitive information from private training dataset, thereby degrading the performance of  MI attack. We conduct an analysis using Fisher Information to justify our method. Our defense is remarkably simple to implement. Without bells and whistles, we show in extensive experiments that TL-DMI achieves state-of-the-art (SOTA) MI robustness. \textbf{Our code, pre-trained models, demo and inverted data are available at: \href{https://hosytuyen.github.io/projects/TL-DMI}{https://hosytuyen.github.io/projects/TL-DMI}}

\end{abstract}    
\section{Introduction}

\begin{figure*}[ht]
  \centering
  \begin{adjustbox}{width=1.0\textwidth,center}
  \includegraphics[width=0.95\textwidth]{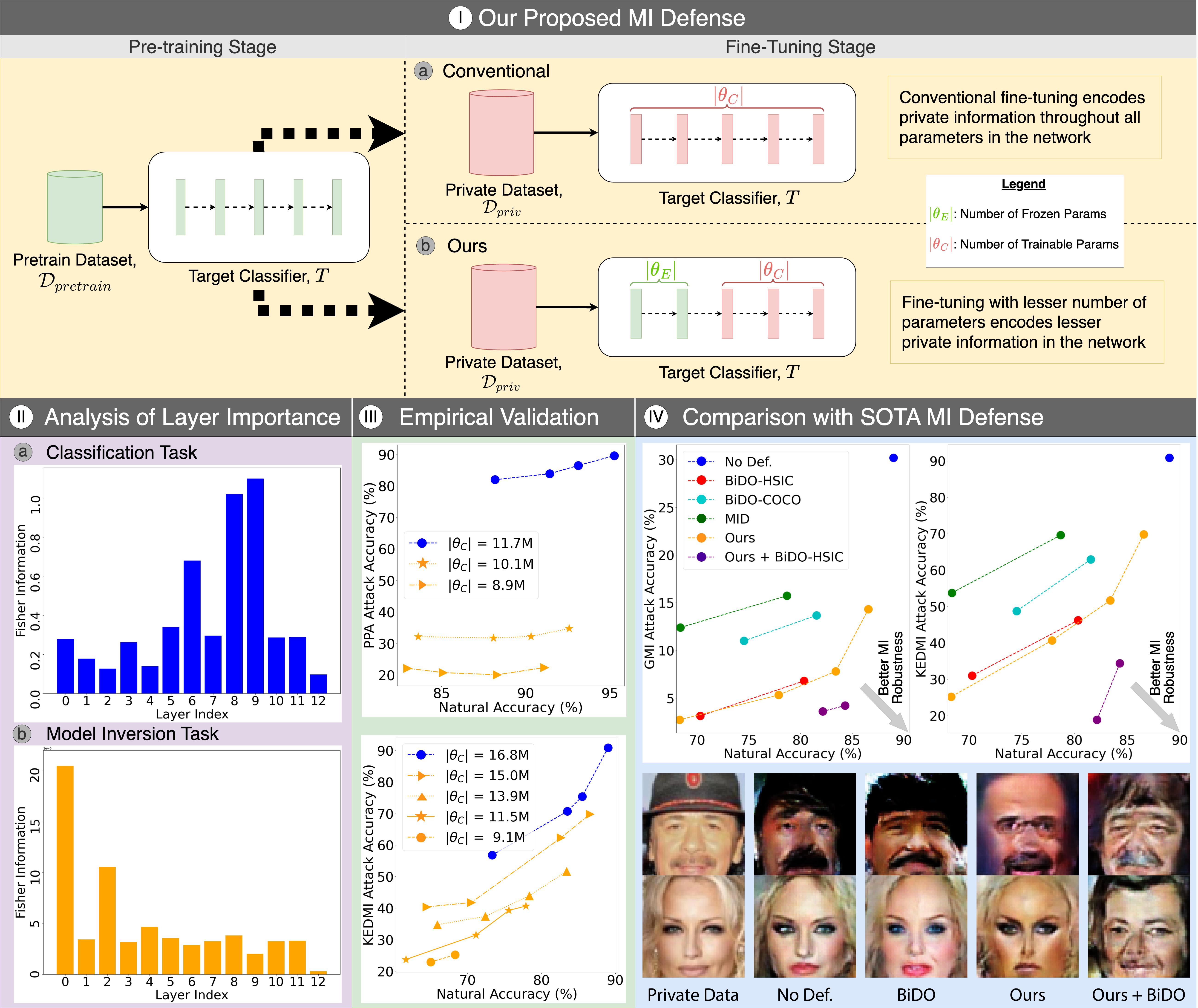}
  \end{adjustbox}
  \label{fig:Overall}
  \vspace{-0.5cm}
  \caption{
  {\bf (I) Our proposed \underline{T}ransfer \underline{L}earning-based \underline{D}efense against \underline{M}odel \underline{I}nversion (TL-DMI) (Sec.~\ref{Proposed Method})}. Based on standard TL framework with pre-training (on public dataset) followed by fine-tuning (on private dataset), we propose a simple and highly-effective method to defend against MI attacks. Our idea is to limit fine-tuning with private dataset to a specific number of layers, thereby limiting the encoding of private information to these layers only (\textcolor{pink}{pink}). Specifically, we propose to perform fine-tuning only on the last several layers. {\bf(II) Analysis of layer importance for classification task and MI task (Sec.~\ref{FI}).} For the first time, we analyze importance of target model layers for MI. For a model trained with conventional training, we apply FI and find that the first few layers of the model are important for MI. Meanwhile, FI analysis suggests that last several layers are important for a specific classification task, consistent with TL literature \citep{yosinski2014transferable}. This supports our hypothesis that preventing the fine-tuning of the first few layers on private dataset could degrade MI significantly, while such impact for classification could be small. Overall, this leads to improved MI robustness. {\bf(III) Empirical validation (Sec.~\ref{Empirical Evidence}).} The sub-figures clearly show that at the same natural accuracy, lower MI attack accuracy can be achieved by reducing the number of parameters fine-tuned with private dataset. {\bf(IV) Comparison with SOTA MI Defense (Sec.~\ref{Comparison with SOTA}).} Without bells and whistles, our method achieves SOTA in MI robustness. Visual quality of MI-reconstructed images from our model is inferior. User study confirms this finding. Extensive experiments can be found in Sec.~\ref{Extensive setups}. \textbf{Best viewed in color with zooming in.}
        }
    \vspace{-0.5cm}
  \label{fig:Overall}
\end{figure*}

Model Inversion (MI) attack is a type of privacy threat that aim to reconstruct private training data by exploiting access to machine learning models. 
State-of-the-art (SOTA)
MI attacks \citep{zhang2020secret, chen2021knowledge, wang2021variational, nguyen_2023_CVPR, nguyen2023labelonly} have demonstrated increased 
% sophistication and 
effectiveness, achieving attack performance of over 90\% in face recognition benchmarks. The implications of this vulnerability are particularly concerning in security-critical applications 
% like face recognition 
\citep{meng2021magface, guo2020learning, huang2020curricularface, schroff2015facenet, dufumier2021contrastive, yang2022towards, dippel2021towards, abdollahzadeh2023survey, chang2020end, krishna2019speech}. 
% dippel2021towards, chang2020end, 
% %, medical diagnosis
% \cite{dufumier2021contrastive, yang2022towards, dippel2021towards}
% %, and speech recognition
% \cite{chang2020end, krishna2019speech}. 

The aim of our work is to propose new perspective to defend against MI attacks and to improve MI robustness. In particular, {\em MI robustness}
pertains to the tradeoff between MI attack accuracy and model utility.
MI robustness involves two critical considerations: Firstly, a MI robust model should demonstrate a significant reduction in  MI attack accuracy, making it difficult for adversaries to reconstruct  private training samples. Secondly,
while defending against MI attacks, the natural accuracy
of a MI robust model 
should remain competitive.
A model with improved MI robustness ensures that it is resilient to 
MI while maintaining its  utility.

{\bf Research gap.}
Despite the growing threat
arising from SOTA MI, 
there are limited studies on defending against MI attacks and improving MI robustness.
Conventionally, 
differential privacy (DP) is used for ensuring the privacy of individuals in datasets. However, DP
has been shown to be ineffective against MI
\citep{fredrikson2014privacy,zhang2020secret,wang2021improving}.
Meanwhile, a few MI defense methods have been proposed.
Particularly,
all existing SOTA MI defense methods are based on the idea of {\em dependency minimization regularization}
\citep{wang2021improving,peng2022bilateral}:
they introduce additional regularization into the training objective, with the goal of minimizing the dependency between input and output/latent representation. The underlying idea of these works 
 is to reduce correlation between input and output/latent, which MI attacks exploit during the inversion. However, reducing correlation between input and output/latent directly undermines accuracy of the model, resulting in considerable degradation in model utility \citep{wang2021improving}. 
To partially restore the model utility, BiDO 
\citep{peng2022bilateral}
proposes to further introduce another regularization to compensate for the reduced correlation between input and latent. However, with two additional regularization along with the original training objective, BiDO requires significant effort in hyperparameter tuning
based on intensive grid search \citep{peng2022bilateral}, 
and is sensitive to small changes in hyperparameters (see our analysis in Supp.)

\textbf{In this paper,} our main hypothesis is that \textit{a model with fewer parameters encoding sensitive information from private training dataset ($\mathcal{D}_{priv}$) could achieve better MI robustness}.
Based on that, we propose a novel \underline{T}ransfer \underline{L}earning-based \underline{D}efense against \underline{M}odel \underline{I}nversion (TL-DMI) (Fig.~\ref{fig:Overall}).
Leveraging on standard two-stages TL framework \citep{pan2010Transfer,yosinski2014transferable}, 
with pre-training on public dataset as the first stage and fine-tuning on private dataset as the second stage, we propose to limit private dataset fine-tuning only on a specific number of layers.
Specifically, in the second stage, we perform private dataset fine-tuning only on the last several layers of the model. 
The first few layers are frozen during the second stage, preventing private information encoded in these layers.
We hypothesize that by 
reducing the number of parameters fine-tuned with private dataset, we could reduce the amount of private information encoded in the model, making it more difficult for adversaries to reconstruct private training data.

To justify our design, we conduct, for the first time, an analysis of model layer importance for the MI task. We propose to apply Fisher Information (FI) to quantify importance of individual layers for MI \citep{kirk2016cata,li2020few}. Our analysis suggests	that first few layers are important for MI. Therefore, by preventing private information encoded in the first few layers as in our proposed method, we could degrade MI significantly. Meanwhile, during pre-training,
 the first few layers learn low level information (edges, colour blobs). It is known that low level information is generalizable across datasets \citep{yosinski2014transferable}. Therefore, our proposed TL-DMI has only small degrade in model utility. Overall, TL-DMI could achieve SOTA MI robustness.
 We remark that TL-DMI is very easy to implement.  In our experiments, we apply TL-DMI to a range of models (CNN, vision transformers), see Sec.~\ref{Extensive setups}.
On the contrary, BiDO has been applied to only VGG16 \cite{simonyan2014very} and ResNet-34 \citep{he2016deep}.
Our contributions are:
\begin{itemize}[leftmargin=0.5cm]

\item
We propose a simple and highly effective \underline{T}ransfer \underline{L}earning-based \underline{D}efense against \underline{M}odel \underline{I}nversion (TL-DMI). 
Our idea is a novel and major departure from existing MI defense based on dependency minimization regularization. Furthermore, while majority of TL work focuses on improving model accuracy \citep{pan2010Transfer,jiang2022transferability}, our work focuses on degrading MI attack accuracy via TL.

\item
We conduct the first study to analyze layer importance for MI task via Fisher Information. Our analysis results suggest that the first few layers are important for MI,  justifying our design to prevent private information encoded in the first few layers.

\item
We conduct empirical analysis to validate that 
lower MI attack accuracy can be achieved by reducing
the number of parameters fine-tuned with private dataset.
Our analysis carefully removes the influence of natural accuracy on MI attack accuracy.

\item We conduct comprehensive experiments to show that our proposed TL-DMI achieves SOTA MI robustness. As TL-DMI is remarkably easy to implement, we extend our experiments for a wide range of model architectures such as vision transformer \citep{tu2022maxvit}, which MI robustness has not been studied before.

\end{itemize}

\section{Background}

The target model $T$ is trained on a private training dataset $\mathcal{D}_{priv} = \{ (x_i, y_i)_{i=1}^N \}$, where $x_i \in \mathbb{R}^{d_X}$ is the facial image and $y_i \in \{0, 1\}^K$ is the identity.
The target classifier $T$ is a $K$-way classifier $T$: $\mathbb{R}^{d_X} \rightarrow \mathbb{R}^K$, with the parameters $\theta_T \in \mathbb{R}^{d_{\theta}}$.

\textbf{Model Inversion (MI) Attack.} 
In MI attacks, an adversary exploits a target model $T$ trained on a private dataset $\mathcal{D}_{priv}$. However, $\mathcal{D}_{priv}$ should not be disclosed. The main goal of MI attacks is to extract information about the private samples in $\mathcal{D}_{priv}$. The existing literature formulates MI attacks as a process of reconstructing an input $\hat{x}$ that $T$ is likely to classify into the preferred class (label) $y$. 
This study primarily focuses on whitebox MI attacks, which are the most dangerous, and can achieve impressive attack accuracy since the adversary has complete access to the target model. For high-dimensional data like facial images, the reconstruction problem is challenging. To mitigate this issue, SOTA MI techniques suggest reducing the exploration area to the meaningful and pertinent images manifold using a GAN. Under white-box MI, the adversary can access $T(\hat{x})$, the $K$-dim vector of soft output, and public dataset $\mathcal{D}_{pub}$ used to train GAN.
The Eq.~\ref{eq:white_box_latent} represents the step of existing SOTA white-box MI attacks \citep{zhang2020secret, chen2021knowledge, an2022mirror, struppek2022plug, nguyen_2023_CVPR}. The details for SOTA MI attacks can be found in the Supp.

\begin{equation}
     w^* = \arg \min_w (- \log P_T (y|G(w)) + \lambda \mathcal{L}_{prior}(w))
     \label{eq:white_box_latent}
\end{equation}
where $-\log P_T (y|G(w))$ denotes identity loss in MI attack, which guides the reconstructed $\hat{x} = G(w)$ that is most likely to be classified as class $y$ by $T$. $G$ refers to generator to generate reconstructed data $\hat{x}$ from latent vector $w$.
The $\mathcal{L}_{prior}$ is the prior loss, which makes use of public information to learn a distributional prior through a GAN. This prior is used to guide the inversion process to reconstruct meaningful images. The hyper-parameter $\lambda$ is to balance prior loss and identity loss.

\begin{table*}[ht]
\centering
\begin{tabular}{lp{4.9cm}p{4.9cm}p{4.9cm}}
\hline
& \centering\textbf{No Defense} & \centering\textbf{Existing MI defenses} & {\quad\quad\textbf{Our proposed TL-DMI}}   \\ \hline
Stage 1 & \multicolumn{3}{c}{Train $T$ with standard objective on $\mathcal{D}_{pretrain}$} \\ \hline
Stage 2 & Fine-tune the whole $T$ with standard objective on $\mathcal{D}_{priv}$ & Fine-tune the whole $T$ with standard objective and additional
{\em dependency minimization 
regularization} on $\mathcal{D}_{priv}$ & Fine-tune only $C$ with standard objective on $\mathcal{D}_{priv}$ \\ \hline
\end{tabular}
\caption{Training procedure for ``no defense'', existing MI defense methods \citep{wang2021improving, peng2022bilateral} and our proposed TL-DMI. Stage 1 (pre-training) is commonly used in existing methods to reduce the requirement for labeled datasets. TL-DMI takes advantage of such setup to defend MI.}
\vspace{-0.4cm}
\label{tab:comparision_method}
\end{table*}

\textbf{Model Inversion (MI) Defense.} In contrast, the MI defense aims at minimizing the disclosure of training samples during the MI optimization process. 
First MI-specific defense strategy is MID \citep{wang2021improving}, which adds a regularization $d(\hat{x}, T(\hat{x}))$ to the main objective during the target classifier's training to penalize the mutual information between inputs $\hat{x}$ and outputs $T(\hat{x})$. Another approach is Bilateral Dependency Optimization (BiDO) \citep{peng2022bilateral}, which minimizes $d(\hat{x}, f)$ to reduce the amount of information about inputs $\hat{x}$ embedded in feature representations $f$, while maximizing $d(f, y)$ to provide $f$ with enough information about $y$ to restore the natural accuracy. {\bf However, both MID and BiDO suffer from the drawback that their regularization, i.e., \boldmath{$d(\hat{x}, T(\hat{x}))$} for MID and \boldmath{$d(\hat{x}, f)$} for BiDO, conflict with the main training objective, resulting in an explicit trade-off between MI robustness and model utility}. BiDO improves this trade-off with $d(f, y)$ but is hyperparameter-sensitive due to the optimization of three objectives, making it difficult to apply. 
% In other words, MID and BiDO reduce MI attack accuracy by suppressing likelihood $P(y|x)$. This leads to an inevitable degrade in classification, where high likelihood $P(y|x)$ is favorable. 

{\bf Model inversion (MI) vs. Membership inference.}
Beside MI, membership inference \cite{rezaei2021difficulty,shafran2021membership,he2022semi,ko2023practical,shokri2017membership} is another privacy attack on machine learning models.
However, {\bf  the focus of our work, i.e., vision MI attacks, is  fundamentally different from membership inference attacks.}
In a membership inference attack, the attacker's objective is to determine whether a specific data point was part of the training dataset used to train the target model. Membership inference attacks are typically formulated as a prediction problem, where an attacker model is trained to output {\em the probability} of a given data point being a member of the training dataset.
In contrast, vision model inversion attacks are usually formulated as an image reconstruction problem. The attacker aims to output {\em the reconstruction} of high-dimensional training images.
While membership inference attacks are limited to determining membership status (in or out of the training dataset) and may not provide fine-grained information about the training data, model inversion attacks attempt to recover the training data itself, which can be more invasive \cite{zhang2020secret}.

% Recent vision model inversion attacks assume the availability of a very generic distributional prior, often modelled by a generative model, to facilitate the reconstruction of high-dimensional images.

\section{\underline{T}ransfer \underline{L}earning-based \underline{D}efense against \underline{M}odel \underline{I}nversion (TL-DMI)} \label{Proposed Method}

\textbf{Transfer Learning (TL).} TL \citep{pan2010Transfer, yin2019feature} is an effective approach to leverage knowledge learned from a general task to enhance performance in a different  task. By performing pre-training on a large general dataset and then fine-tuning on a  target dataset, TL mitigates the demand for large labeled datasets, while simultaneously improving generalization and overall performance. In machine learning, TL works mostly focus on  improving the model performance by adapting the knowledge to new tasks and domains \citep{jiang2022transferability, zhuang2020comprehensive}.

\noindent \textbf{Our proposed defense TL-DMI.} In contrast,  our work  is the first to apply TL to defend against MI attacks aiming at degrading  MI attack accuracy. Therefore, our study is fundamentally different from existing TL works which aim to improve model utility \citep{pan2010Transfer, yang2019neural, kumar2022fine, kamath2019transfer, kolesnikov2019large}.
Our idea is to apply TL to reduce the leak of private information  by limiting the number of parameters updated on private training data.
Specifically, as illustrate in Fig.~\ref{fig:Overall},
we propose to train the target model $T$ as $T = C \circ E$ in two stages: pre-training and then fine-tuning. Particularly, in the fine-tuning stage, 
$E$ comprises parameters that are frozen, i.e., not updated by the private dataset $\mathcal{D}_{priv}$, while $C$ comprises parameters that are updated by $\mathcal{D}_{priv}$. 

\begin{itemize}[leftmargin=0.5cm]
  \item \textbf{Stage 1: Pre-training with \boldmath{$\mathcal{D}_{pretrain}$.}}
    We first pre-train $T$ using a dataset $\mathcal{D}_{pretrain}$. $\mathcal{D}_{pretrain}$ can be a general domain dataset, e.g., Imagenet1K, or it can be similar  domain as the private dataset $\mathcal{D}_{priv}$.  Importantly, $\mathcal{D}_{pretrain}$ has no class/identity intersection with $\mathcal{D}_{priv}$. Both $C$ and $E$ are updated based on $\mathcal{D}_{pretrain}$ in this stage.

  \item \textbf{Stage 2: Fine-tuning with \boldmath{$\mathcal{D}_{priv}$.}}
    To adapt the pre-trained model from Stage 1 for $\mathcal{D}_{priv}$, we freeze $E$, i.e. parameters of $E$ are unchanged. We only update $C$ with $\mathcal{D}_{priv}$.
\end{itemize}

Tab.~\ref{tab:comparision_method} provides a comparison between our defense TL-DMI and existing MI defenses.  {\em We remark that pre-training has already been commonly adopted in previous works of MI attack. Therefore, in many cases, our method does not incur additional overhead \citep{peng2022bilateral, nguyen_2023_CVPR, chen2021knowledge, struppek2022plug, an2022mirror}.}
As an example, we consider the main setup of BiDO \cite{peng2022bilateral} where  VGG16 \cite{simonyan2014very} is used  as the target classifier $T$. Following the previous works on MI attack, $T$ including $E$ and $C$ are first pre-trained on  $\mathcal{D}_{pretrain}$ = Imagenet1K \citep{deng2009imagenet}. 
Then, for TL-DMI, we fine-tune $C$ with $\mathcal{D}_{priv}$ = CelebA \citep{liu2015deep} while $E$ is frozen. In contrast, for other MI defense, both $E$ and $C$ are updated with $\mathcal{D}_{priv}$. We explore the design of $T$
with different number of layers updated by $\mathcal{D}_{priv}$, leading to  different number of parameters in $C$  ($|\theta_C|$) updated by $\mathcal{D}_{priv}$. 
Using different $|\theta_C|$, we limit the amount of private information encoded in the parameters of $T$. We show that our approach TL-DMI improves  MI robustness.

Regarding  hyperparameter in our proposed TL-DMI, we determine  $|\theta_C|$ by simply deciding at the {\em layer-level} of a deep neural network. Note that during training we use the same objective of classification task, i.e. no change in training objective is needed. Therefore, TL-DMI is much simpler and faster than SOTA MI defense BiDO \citep{peng2022bilateral} (see Supp.).
{\bf In Sec. \ref{FI}, we present our Fisher Information-based analysis to justify TL-DMI.}

\section{Exploring MI Robustness via Transfer Learning} \label{Sec:Study}

We introduce the experiment setup in Sec.~\ref{sec:Experimental_Setup}. In Sec.~\ref{FI}, we provide the first analysis on layer importance for MI task via Fisher Information suggesting that earlier layers are important for MI. Then, Sec.~\ref{Empirical Evidence} empirically validate that MI robustness is obtained by reducing the number of parameters fine-tuned with private dataset. With the established understandings, we then compare our proposed method with current SOTA MI defenses \citep{wang2021improving, peng2022bilateral} in Sec.~\ref{Comparison with SOTA}. Additionally, since our method offer higher practicality compared with the SOTA MI defenses, we extensively access our approach on 20 MI attack setups in Sec.~\ref{Extensive setups} and Supp.,  spanning 9 architectures, 4 private datasets $\mathcal{D}_{priv}$, 3 public datasets $\mathcal{D}_{pub}$, and 7 MI attacks.

While the above sections assume a consistent pre-trained dataset $\mathcal{D}_{pretrain}$ for the target classifier to ensure fair comparison with existing works, we also delve into novel analysis on the effect of various $\mathcal{D}_{pretrain}$ on MI robustness. We observe that \textit{less similarity between pretrain and private dataset domains can improve defense effectiveness}. The details for this analysis can be found in Supp.  

\subsection{Experimental Setup} \label{sec:Experimental_Setup}

\textbf{To ensure a fair comparison, our study strictly follows setups in SOTA MI defense method BiDO \citep{peng2022bilateral} in datasets, attack methods, and network architectures.} Furthermore, we also examine our defense approach with additional new datasets, recent MI attack models, and new network architectures. Note that these have not been included in BiDO. All the MI setups in our study are summarized in Tab.~\ref{tab:Setup}. The details for the setup can be found in Supp.

\textbf{MI Defense Baseline.} In order to showcase the efficacy of our proposed TL-DMI, we compare TL-DMI with several existing SOTA model inversion defense methods, which are BiDO \citep{peng2022bilateral} and MID \citep{wang2021improving}.

\textbf{Evaluation Metrics.} Following the previous MI defense/attack works, we adopt  natural accuracy (Acc), Attack Accuracy (AttAcc), K-Nearest Neighbors Distance (KNN Dist), and $\ell_2$ distance metrics to evaluate MI robustness. Moreover, we also provide qualitative results and user study in the Supp.

\begin{table}[h]
\setlength{\tabcolsep}{2pt}
\begin{adjustbox}{width=1.0\columnwidth}
\begin{tabular}{p{2.4cm}p{1.9cm}p{2.5cm}p{2.55cm}}
\hline
\textbf{Attack Method}        \T\B          & \boldmath{$\mathcal{D}_{pub}$}     & \boldmath{$\mathcal{D}_{priv}$}    & \boldmath{$T$}                    \\ \hline
VMI    \cite{wang2021variational} \T\B                   & \multirow{5}{*}{CelebA \cite{liu2015deep}} & \multirow{5}{*}{CelebA \cite{liu2015deep}} & ResNet-34  \cite{he2016deep}            \\ \cline{1-1} \cline{4-4} 
\T KEDMI\cite{chen2021knowledge}/ GMI \cite{zhang2020secret}      \B            &                         &                         & \multirow{4}{*}{VGG16 \cite{simonyan2014very}} \\ \cline{1-1}

\T LOMMA \cite{nguyen_2023_CVPR} / BREPMI \cite{kahla2022label}              &                         &                         &                        \\ \hline
\multirow{2}{*}{KEDMI \cite{chen2021knowledge} / } & \T CelebA \cite{liu2015deep} / FFHQ \B  \cite{karras2019style}         & \multirow{3}{*}{CelebA \cite{liu2015deep}} & FaceNet64\cite{cheng2017know}/ IR152 \cite{he2016deep}         \\ \cline{2-2} \cline{4-4} 
   \vspace{-5mm}    GMI \cite{zhang2020secret}                    & \T\B FFHQ   \cite{karras2019style}                 &                         & VGG16      \cite{simonyan2014very}            \\ \hline
\multirow{4}{*}{PPA \cite{struppek2022plug}}       & \vspace{2mm} FFHQ    \cite{karras2019style}                & \vspace{2mm} FaceScrub \cite{ng2014data}              & \T ResNet-18 \cite{he2016deep} / ResNet-101 \cite{he2016deep} / MaxViT \cite{tu2022maxvit} \B   \\ \cline{2-4} 
                           & AFHQ   \cite{choi2020stargan}                 & StanfordDogs  \T \B  \cite{dataset2011novel}        & ResNeSt-101    \cite{zhang2022resnest}        \\ \hline
MIRROR  \cite{an2022mirror}                  & FFHQ  \T\B  \cite{karras2019style}                & VGGFace2 \cite{cao2018vggface2}               & ResNet-50   \cite{he2016deep}           \\ \hline
\end{tabular}
\end{adjustbox}
\caption{{\bf Setups of our comprehensive experiments.} We follow the exact setups in the previous MI attacks. Beside the standard MI setups on GMI \cite{zhang2020secret}/KEDMI \cite{chen2021knowledge} on VGG16, and VMI \cite{wang2021variational} on Resnet-34, we also evaluate our defense approach on current SOTA MI setups. Due to the need of intensive grid-search for hyper-paramters, it is very time consuming to expand the exisiting SOTA MI Defense \cite{peng2022bilateral} to these additional MI setups. In total, there are 20 MI setups spanning 7 MI attacks, 3 $\mathcal{D}_{pub}$, 4 $\mathcal{D}_{priv}$, 9 architectures of $T$. The experimental setups are described in more detail in the Supp.}
\vspace{-0.3cm}
\label{tab:Setup}
\end{table}

\subsection{Analysis of Layer Importance for
Classification Task and MI Task} \label{FI}

In this section, we provide an analysis to
justify our proposed TL-DMI to render MI robustness. 
We aim to understand importance of individual layers for MI reconstruction task, justifying our design in TL-DMI to prevent encoding of private data information in the first few layers as an effective method to degrade MI. 
We study layer importance between  classification and MI tasks. To quantify the importance, we compute the Fisher Information (FI) for the two tasks for individual layers. 

\textbf{Fisher Information (FI) based analysis.} 
Fisher Information $F$ has been applied to measure the importance of model parameters
for discriminative task
\citep{kirk2016cata,achille2019task2vec}
and generative task 
\citep{li2020few}. 
For example, in \citep{kirk2016cata}, FI has been applied to determine the importance of model parameters to overcome the catastrophic forgetting in continual learning.
Our study extends FI-based analysis for model inversion, which has not been studied before.
Specifically, 
given a model $T$ parameterized by $\theta_T$ and input $X$,  FI can be computed as
\citep{kirk2016cata,achille2019task2vec,li2020few}:
\begin{equation}
    F = \mathbb{E} \left[ -\frac{\partial^2}{\partial\theta_T^2}\mathcal{L}(X|\theta_T) \right]
    \label{eq:general_FI}
\end{equation}
Here, $\mathcal{L}$ is the loss function for a particular task. Specifically, we investigate  FI on classification task and MI task.
For classification, we follow  \citet{achille2019task2vec} and \citet{le2021fisher} to use cross entropy $\mathbb{E} \left[ - \log p(y_i|x_i) \right]$ as $\mathcal{L}$ and validation set $\mathcal{D}_{priv}^{val} = \{ (x_i, y_i)_{i=1}^M \}$ as $X$. For MI task, we propose to use the $\ell_2$ distance between the feature representations of reconstructed images and the private images as $\mathcal{L}$: 
\begin{equation}
    \mathbb{E} \left[ \Bigl\lVert \Phi(\hat{x}_{{u}}^j) - \mathbb{E} \left[ \Phi(x_{priv}^j) \right] \Bigl\lVert_{2} \right]
\end{equation}
Here, for a given input image, $\Phi$ computes the penultimate layer representation using the target model, and 
 $\hat{x}_{{u}}^j$ is one of the  MI reconstructed images for identity $j$, and 
$\mathbb{E} \left[ \Phi(x_{priv}^j) \right]$ is the centroid feature of private images for identity $j$. Therefore, we use the distance between MI reconstructed image and private image of the same identity as the loss in FI analysis. The set of MI reconstructed images $ \{  \hat{x}_{{u}}^j  \}_{j=1}^{J}$ for different identity is used as $X$. We explore different setups to compute $\mathcal{L}$, see Supp.  In one setup, we perform FI analysis only at the last iteration (i.e., 3000, for the result in Fig.~\ref{fig:Overall}-II). As we are interested in FI at the layer level, we compute the average FI of all parameters within a layer. We use  the main MI attack setup in \citet{peng2022bilateral}, i.e.,  VGG16 with KEDMI \cite{chen2021knowledge} attack, for FI analysis.

\textbf{Observation.} The FI results in Fig.~\ref{fig:Overall}-II clearly suggest that the first few layers of a target model are important for MI task.
Meanwhile, FI analysis suggests that the first few layers do not carry important information for a specific classification task.
This observation is consistent with previous finding in work \citep{yosinski2014transferable} suggesting that the earlier layers
carry general features. 
The FI analysis justifies our design to prevent encoding of private information in the 
first few layers in order to degrade MI attacks, while keeping the impact on classification small.
Overall, this leads to improved MI robustness.
{\bf 
Further results with different loss (\boldmath{$\ell_1$} and LPIPS \citep{zhang2018unreasonable}) 
and different MI iterations can be found in Supp.}

\subsection{Empirical Validation} \label{Empirical Evidence}

As shown in Fig.~\ref{fig:Overall}-IV, we observe a significant improvement in MI robustness when reducing the number of parameters fine-tuned with $\mathcal{D}_{priv}$. However, the relationship between MI attack accuracy and natural accuracy is strongly correlated \citep{zhang2020secret}, which makes it unclear if the decrease in MI attack accuracy is due to the drop in natural accuracy. 

In this section, we empirically investigate the hypothesis that \textit{a model with fewer parameters encoding private information from $\mathcal{D}_{priv}$ has better MI robustness}. The empirical validation is reported in Fig.~\ref{fig:Overall}-III. Note that the number of parameters for the entire target model:  {\color{blue} $|\theta_C|$ = 16.8M} for VGG16 with KEDMI \cite{chen2021knowledge} setup and {\color{blue} $|\theta_C|$ = 11.7M} for Resnet-18 with PPA \cite{struppek2022plug} setup. The additional empirical validation for GMI can be found in the Supp. To separate the influence of model accuracy on MI attack accuracy, we perform PPA/KEDMI attacks on different checkpoints for each training setup, varying a wide range of natural accuracy. This is presented by multiple data points on each line.

The results clearly show that fine-tuning fewer parameters on $\mathcal{D}_{priv}$ enhances MI robustness compared with fine-tuning all parameters on $\mathcal{D}_{priv}$, regardless of the effect on natural accuracy. For instance, in the KEDMI setup, with a comparable natural accuracy of 83\%, fine-tuning only {\color{orange} $|\theta_C|$ = 13.9M} reduces a third attack accuracy compared to fine-tuning {\color{blue} $|\theta_C|$ = 16.8M}. The result in the PPA setup is even more supportive, where with a natural accuracy of around 91\%, fine-tuning {\color{orange} $|\theta_C|$ = 8.9M} reduces the attack accuracy to 22.36\% from 91.7\% in {\color{blue} $|\theta_C|$ = 11.7M}.

Across all configurations, we observe that the fewer parameters fine-tuned on $\mathcal{D}_{priv}$, the more robust the model. However, it is important to note that if the number of fine-tuned parameters on $\mathcal{D}_{priv}$ is insufficient, such as $|\theta_C|$ = 9.1M for KEDMI setup, the model's natural accuracy may drop drastically, rendering it unusable. Overall, our experiments strongly suggest  that  \textbf{better MI robustness can be achieved by reducing the number of parameters fine-tuned on \boldmath{${\mathcal{D}_{priv}}$}}.

\subsection{Comparison with SOTA MI Defense}
\label{Comparison with SOTA}

In this section, we compare our proposed TL-DMI defense with current existing MI defenses \cite{peng2022bilateral, wang2021improving}.
\textit{For a fair comparison, we strictly follow the setups in SOTA MI defense \citep{peng2022bilateral}}. Specifically, we first present the MI robustness comparison against KEDMI/GMI in Fig.~\ref{fig:Overall}-IV. MID \cite{wang2021improving} improves MI robustness by penalizing the mutual information between inputs and outputs during the training process, which is intractable in continuous and high-dimensional settings, making MID resort to mutual information approximations rather than actual quantity \citep{peng2022bilateral}.
In general, MID is outperformed by more recent defense BiDO \cite{peng2022bilateral}.

% Therefore, we need to sacrifice significant model accuracy to observe the improvement in MI robustness, which results in a poor MI robustness compared with the SOTA defense BiDO \cite{peng2022bilateral}. 

Our proposed TL-DMI is simple yet effective, achieving outstanding MI robustness as shown in
Fig.~\ref{fig:Overall}-IV.
% than BiDO without requiring additional conflict regularization, making it more feasible to recover model accuracy. 
We are the 
first to explore MI defense beyond the regularization perspective. TL-DMI can be combined with SOTA MI defenses such as BiDO. When combining with TL-DMI, we strictly follow BiDO. The only difference is that BiDO is applied only to the unfrozen layers in the fine-tuning stage. The results in Fig.~\ref{fig:Overall}-IV show that the trade-off between utility and robustness is much improved when we combine two approaches. Also, TL-DMI helps restore the utility degraded by BiDO, rendering a much more MI robust model (reducing MI attack accuracy by 27.36\% from 46.23\% to 18.87\%) while improving model utility (increasing model accuracy by 1.8\% from 80.35\% to 82.15\%). 

In VMI setup presented in Tab.~\ref{tab:comparison_SOTA}, MID \cite{wang2021improving} suffers when applied to VMI \cite{wang2021variational} due to the requirement of modifying the last layer of the network to implement the variational approximation of the mutual information \cite{peng2022bilateral}. Hence, we observe a significant drop in natural accuracy when applying MID \cite{wang2021improving} to VMI \cite{wang2021variational}. BiDO \cite{peng2022bilateral} partially addresses this problem and recovers natural accuracy better with comparable attack accuracy. Comapred to BiDO, TL-DMI updating $|\theta_C|$ = 21.14M (out of 21.5M parameters in totam) improves natural accuracy by around 1\%-3\% while achieving greater robustness by reducing attack accuracy by around 6\%. 

% \usepackage{multirow}
\begin{table}[]
\begin{adjustbox}{width=1.0\columnwidth,center}
\begin{tabular}{cclccc}
\hline
\textbf{Attack Method} \T\B           & \boldmath{$T$}                         & \textbf{Defense}                 & \textbf{Acc} \boldmath{$\Uparrow$}   & \textbf{AttAcc} \boldmath{$\Downarrow$} &  \boldmath{$\Delta \Uparrow$}      \\ \hline
\multirow{3}{*}{VMI \cite{wang2021variational}}  & \multirow{3}{*}{ResNet-34 \cite{he2016deep}}  & No Def.      \T           & 69.27 & 39.40  & -              \\
                       &                             & BiDO                    & 61.14 & 30.25  & 1.13           \\
                       &                             & TL-DMI  \B & 62.20 & 23.70  & \textbf{2.22}  \\ \hline
\multirow{4}{*}{LOMMA \cite{nguyen_2023_CVPR}}  & \multirow{4}{*}{VGG-16 \cite{simonyan2014very}}     & No Def.        \T          & 89.00 & 95.67  & -              \\
                       &                             & BiDO                    & 80.35 & 70.47  & 2.91           \\
                       &                             & TL-DMI  & 83.41 & 75.67  & \textbf{3.58}  \\
                       % &                             & TL-DMI -b \B & 78.86 & 59.68  & \textbf{3.54}  \\ 
                       \hline
\multirow{6}{*}{PPA \cite{struppek2022plug}}   & \multirow{3}{*}{ResNet-18 \cite{he2016deep}}  & No Def.                  & 94.22 & 88.46  & -              \\
                       &                             & BiDO                    & 91.33 & 76.56  & 4.12           \\
                       &                             & TL-DMI  & 91.12 & 22.36  & \textbf{21.32} \\ \cline{2-6} 
                       & \multirow{3}{*}{ResNet-101 \cite{he2016deep}} & No Def.                  & 94.86 & 83.00  & -              \\
                       &                             & BiDO                    & 90.31 & 67.26  & 3.46           \\
                       &                             & TL-DMI  & 90.10 & 31.82  & \textbf{10.75} \\ \hline
\end{tabular}
\end{adjustbox}
\caption{The comparison between our proposed TL-DMI and SOTA MI denfense BiDO \cite{peng2022bilateral}, where the Acc and AttAcc are given in \%.
Our  evaluation covers a wide range of MI attack setups. We follow previous work for MI setups (see details in Tab.~\ref{tab:Setup} and Supp.). To implement TL-DMI, we set $|\theta_C|$ = 21.14M/13.90M/8.90M/16.05M for $T$ = ResNet-34/VGG-16/ResNet-18/ResNet-101, respectively. {\bf MI robustness is quantified by the $\Delta$, the ratio of drop in attack accuracy to drop in natural accuracy.} 
As shown in the results, 
our proposed TL-DMI significantly improves 
MI robustness comparing to BiDO. 
}
\vspace{-0.5cm}
\label{tab:comparison_SOTA}
\end{table}

In another effort to comprehensively compare with the SOTA MI defense BiDO \cite{peng2022bilateral}, we extend the evaluation to include additional SOTA MI attacks: LOMMA \cite{nguyen_2023_CVPR} and PPA \cite{struppek2022plug}. Given the different setup of PPA compared to BiDO, we adapt BiDO to work with additional architectures of $T$, specifically ResNet-18/101 \cite{he2016deep}, tailored to the PPA attack. Note that these evaluations have not been explored yet in the MI literature \cite{peng2022bilateral,wang2021improving}. From the results in Tab.~\ref{tab:comparison_SOTA}, a consistent trend is that all defenses have suffered in natural accuracy, but TL-DMI method has suffered the least in natural accuracy while reducing the most in attack accuracy. Consequently, TL-DMI achieves the best MI robustness trade-off, which can be quantified by $\Delta$, which is the ratio of drop in attack accuracy to drop in natural accuracy (the larger is the ratio, the better is MI robustness). Additional comparison against BREPMI \citep{kahla2022label} can be found in Supp. In conclusion, our proposed TL-DMI stands out as highly effective across a range of SOTA MI attacks \cite{struppek2022plug,nguyen_2023_CVPR}. 

\subsection{Extended MI Robustness Evaluation} \label{Extensive setups}

\begin{table*}[]
\setlength{\tabcolsep}{2pt}

\begin{adjustbox}{width=1.80\columnwidth,center}
\begin{tabular}{>{\centering\arraybackslash}p{1.5cm}ccccp{1.4cm}p{1.6cm}>{\centering\arraybackslash}p{1.0cm}>{\centering\arraybackslash}p{2.1cm}>{\centering\arraybackslash}p{2.1cm}p{0.9cm}}
\hline
\textbf{Attack Method}     & \boldmath{$\mathcal{D}_{priv}$}     & \boldmath{$\mathcal{D}_{pub}$}      & \boldmath{$\mathcal{D}_{pretrain}$}         & \boldmath{$T$}  \T\B       & \textbf{Defense Method} & \textbf{\boldmath{$|\theta_C| \slash |\theta_T|$}} \T\B & \textbf{Acc \quad $\Uparrow$} & \textbf{Top1-AttAcc $\Downarrow$} & \textbf{Top5-AttAcc $\Downarrow$} & \textbf{KNN Dist} $\Uparrow$ \\ \hline
\multirow{12}{*}{KEDMI} & \multirow{6}{*}{CelebA} & \multirow{6}{*}{CelebA} & \multirow{2}{*}{ImageNet1K}  & \multirow{2}{*}{VGG16}   & No Def. \T\B         & 16.8/16.8            & 89.00           & 90.87 $\pm$ 2.71          & 99.33 $\pm$ 0.75          & 1168       \\
            &             &             &                &              & TL-DMI           & 13.9/16.8            & 83.41           & \textbf{51.67 $\pm$ 3.93}      & \textbf{80.33 $\pm$ 2.91}      & \textbf{1410}   \\ \cline{4-11} 
            &             &             & \multirow{4}{*}{MS-CelebA-1M} & \multirow{2}{*}{IR152}   & No Def. \T\B         & 62.6/62.6            & 93.52           & 94.07 $\pm$ 1.82          & 99.67 $\pm$ 0.63          & 1071       \\
            &             &             &                &              & TL-DMI           & 17.8/62.6            & 86.70           & \textbf{64.60 $\pm$ 4.93}      & \textbf{87.67 $\pm$ 2.73}      & \textbf{1333}   \\ \cline{5-11} 
            &             &             &                & \multirow{2}{*}{FaceNet64} & No Def. \T\B         & 35.4/35.4            & 88.50           & 86.73 $\pm$ 2.85          & 98.33 $\pm$ 1.49          & 1194       \\
            &             &             &                &              & TL-DMI           & 34.4/35.4                &    83.41         & \textbf{73.40 $\pm$ 4.10}               & \textbf{91.67 $\pm$ 1.92}               & \textbf{1265}        \\ \cline{2-11} 
            & \multirow{6}{*}{CelebA} & \multirow{6}{*}{FFHQ}  & \multirow{2}{*}{ImageNet1K}  & \multirow{2}{*}{VGG16}   & No Def. \T\B         & 16.8/16.8            & 89.00           & 55.60 $\pm$ 3.75          & 84.67 $\pm$ 2.85          & 1407       \\
            &             &             &                &              & TL-DMI           & 13.9/16.8            & 83.41           & \textbf{34.53 $\pm$ 3.43}      & \textbf{65.33 $\pm$ 3.36}      & \textbf{1554}   \\ \cline{4-11} 
            &             &             & \multirow{4}{*}{MS-CelebA-1M} & \multirow{2}{*}{IR152}   & No Def. \T\B         & 62.6/62.6            & 93.52           & 70.27 $\pm$ 3.40          & 89.33 $\pm$ 2.14          & 1285       \\
            &             &             &                &              & TL-DMI           & 17.8/62.6            & 86.70           & \textbf{46.53 $\pm$ 4.58}      & \textbf{72.67 $\pm$ 3.16}      & \textbf{1454}   \\ \cline{5-11} 
            &             &             &                & \multirow{2}{*}{FaceNet64} & No Def. \T\B         & 35.4/35.4            & 88.50           & 57.87 $\pm$ 4.70          & 82.00 $\pm$ 3.45          & 1409       \\
            &             &             &                &              & TL-DMI           & 34.4/35.4                &    83.41            & \textbf{15.27 $\pm$ 4.09}               & \textbf{31.00 $\pm$ 4.24}               & \textbf{1751}        \\ \hline
\multirow{12}{*}{GMI}  & \multirow{6}{*}{CelebA} & \multirow{6}{*}{CelebA} & \multirow{2}{*}{ImageNet1K}  & \multirow{2}{*}{VGG16}   & No Def. \T\B         & 16.8/16.8            & 89.00           & 30.20 $\pm$ 5.26          & 55.00 $\pm$ 5.95          & 1600       \\
            &             &             &                &              & TL-DMI           & 13.9/16.8            & 83.41           & \textbf{7.80 $\pm$ 3.36}      & \textbf{23.33 $\pm$ 4.60}      & \textbf{1845}   \\ \cline{4-11} 
            &             &             & \multirow{4}{*}{MS-CelebA-1M} & \multirow{2}{*}{IR152}   & No Def. \T\B         & 62.6/62.6            & 93.52           & 40.87 $\pm$ 4.76          & 66.67 $\pm$ 5.76          & 1516       \\
            &             &             &                &              & TL-DMI           & 17.8/62.6            & 86.70           & \textbf{8.93 $\pm$ 3.73}      & \textbf{22.67 $\pm$ 5.21}      & \textbf{1819}   \\ \cline{5-11} 
            &             &             &                & \multirow{2}{*}{FaceNet64} & No Def. \T\B         & 35.4/35.4            & 88.50           & 26.87 $\pm$ 3.75          & 49.00 $\pm$ 6.05          & 1643       \\
            &             &             &                &              & TL-DMI           & 34.4/35.4     &   83.61            & \textbf{15.73 $\pm$ 4.58}               & \textbf{33.00 $\pm$ 6.28}               & \textbf{1752}        \\ \cline{2-11} 
            & \multirow{6}{*}{CelebA} & \multirow{6}{*}{FFHQ}  & \multirow{2}{*}{ImageNet1K}  & \multirow{2}{*}{VGG16}   & No Def. \T\B         & 16.8/16.8            & 89.00           & 13.60 $\pm$ 4.43          & 32.00 $\pm$ 4.92          & 1725       \\
            &             &             &                &              & TL-DMI           & 13.9/16.8            & 83.41           & \textbf{4.27 $\pm$ 2.56}      & \textbf{12.33 $\pm$ 3.44}      & \textbf{1919}   \\ \cline{4-11} 
            &             &             & \multirow{4}{*}{MS-CelebA-1M} & \multirow{2}{*}{IR152}   & No Def. \T\B         & 62.6/62.6            & 93.52           & 24.27 $\pm$ 4.24          & 45.67 $\pm$ 6.71          & 1617       \\
            &             &             &                &              & TL-DMI           & 17.8/62.6            & 86.70           & \textbf{6.13 $\pm$ 3.11}      & \textbf{15.00 $\pm$ 4.98}      & \textbf{1877}   \\ \cline{5-11} 
            &             &             &                & \multirow{2}{*}{FaceNet64} & No Def. \T\B         & 35.4/35.4            & 88.50           & 13.13 $\pm$ 4.96          & 30.33 $\pm$ 5.40          & 1746       \\
            &             &             &                &              & TL-DMI           & 34.4/35.4     &   83.61        & \textbf{2.60 $\pm$ 1.49}            & \textbf{8.67 $\pm$ 3.64}               & \textbf{2009}                      \\ \hline

\end{tabular}
\end{adjustbox}
\caption{Our evaluation covers multiple MI attack setups, target models, and public, private and pre-trained datasets.  Here, 
 the results are given in \%. Specifically, we reports the MI defense results against different MI attack methods (KEDMI and GMI), as well as using different public datasets $\mathcal{D}_{pub}$ (CelebA and FFHQ), and pre-trained datasets $\mathcal{D}_{pretrain}$ (Imagenet1K and MS-CelebA-1M), for several target model $T$: VGG16, IR152, FaceNet64.}
 \vspace{-0.2cm}
\label{tab:Extensive_setup}
\end{table*}

Our proposed TL-DMI is simple, easy to implement, and less sensitive to hyperparameters than BiDO, which requires intensive grid search for hyperparameter. This significant advantage allows us to extend the scope of experimental setups for the MI defense to align with the remarkable increase in MI attack setups, which are not yet evaluated in previous MI defenses \citep{peng2022bilateral, wang2021improving}. 

% Due to the notably high time demands associated with hyper-parameter search for BiDO, we have opted to exclude BiDO results in this extended evaluation.

\textbf{Results on different \boldmath{${\mathcal{D}_{pub}}$}}. We evaluate TL-DMI against KEDMI and GMI attacks on three architectures (VGG16, IR152, FaceNet64) with varying public datasets (CelebA, FFHQ), spanning 12 facial domain MI setups. These are standard setups in KEDMI/GMI, however, only 2 out of 12 setups examined in the current SOTA MI defense were presented in \citep{peng2022bilateral}. The results in Tab.~\ref{tab:Extensive_setup} demonstrate that TL-DMI consistently achieves significantly more robust models across all setups while maintaining acceptable natural accuracy, with significant improvements in robustness across a wide range of attack scenarios (13.33\%-42.60\% for KEDMI, 11.14\%-31.94\% for GMI). \textit{On average, TL-DMI  significantly reduces the accuracy of MI attacks by more than a half.}

\textbf{Results on SOTA high resolution MI attacks}. Furthermore, we provide our defense results against SOTA High Resolution MI attacks, i.e., PPA \citep{struppek2022plug} and MIRROR \cite{an2022mirror} in Tab.~\ref{tab:comparison_SOTA} and Tab.~\ref{tab:SOTA_highres}. To the best of our knowledge, our work is the first MI defense approach against such high resolution MI attack. The results are very encouraging. We observe only a small reduction in natural accuracy, while the attack accuracy experiences a significant drop thanks to our defense TL-DMI.

{\begin{table*}[]
\centering

\tiny
\begin{adjustbox}{width=1.80\columnwidth,center}
\begin{tabular}{p{0.6cm}p{1cm}p{0.8cm}p{0.7cm}>{\centering\arraybackslash}p{0.25cm}>{\centering\arraybackslash}p{0.3cm}>{\centering\arraybackslash}p{0.5cm}>{\centering\arraybackslash}p{0.9cm}>{\centering\arraybackslash}p{0.7cm}>{\centering\arraybackslash}p{0.35cm}}
\hline
\T \textbf{Attack Method}          & \boldmath{$\mathcal{D}_{priv}$}                    & \boldmath{$T$}                          & \textbf{Defense}       & \textbf{Acc} \boldmath{$\Uparrow$}    & \textbf{AttAcc}  \boldmath{$\Downarrow$}   & \boldmath{$\delta_{Eval}$} \boldmath{$\Uparrow$}      \B      & \boldmath{$\delta_{FaceNet}$} \boldmath{$\Uparrow$}         & \boldmath{$\ell_2$} \textbf{Dist} \boldmath{$\Uparrow$}              & \textbf{FID} \boldmath{$\Uparrow$}            \\ \hline
\multirow{4}{*}{PPA}    &     \multirow{2}{*}{FaceScrub}                  & \multirow{2}{*}{MaxViT}      &\T  No Def.       & 96.57          & 79.63          & 128.46          & 0.7775          & -               & 50.37          \\
                        &                                &                              & TL-DMI \B & 93.01 & \textbf{21.17} & \textbf{168.85} & \textbf{1.0199} & \textbf{-}      & \textbf{55.50} \\ \cline{2-10} 
                        & \multirow{2}{*}{Stanford Dogs} & \multirow{2}{*}{ResNeSt-101} &\T  No Def.       & 75.07          & 91.90          & 62.56           & -               & -               & 33.69          \\
                        &                                &                              & TL-DMI  \B & 79.54 & \textbf{60.88} & \textbf{83.57}  & \textbf{-}      & \textbf{-}      & \textbf{46.01} \\ \hline
\multirow{2}{*}{MIRROR} & \multirow{2}{*}{VGGFace2}      & \multirow{2}{*}{ResNet-50}   & \T No Def.      & 99.44          & 84.00          & -               & -               & 602.41          & -              \\
                        &                                &                              & TL-DMI  \B & 99.40 & \textbf{50.00} & \textbf{-}      & \textbf{-}      & \textbf{650.28} & \textbf{-}     \\ \hline
\end{tabular}
\end{adjustbox}
\caption{The defense results for SOTA MI attacks on 224x224 images. We strictly follow experimental setups from PPA and MIRROR, presenting results for Acc and AttAcc in \%. Additionally, we employ PPA-introduced metrics, $\delta_{FaceNet}$ and $\delta_{Eval}$, alongside MIRROR-introduced metric $l_2$ Dist for the evaluation. 
Our proposed TL-DMI successfully defends against SOTA MI attacks on high resolution 224x224. To train our TL-DMI defense models, we set $|\theta_C|$ = 18.3M/27.9M/32.9M for $T$ = MaxViT/ResNeSt-101/ResNet-50, respectively.}
\label{tab:SOTA_highres}
\vspace{-5mm}
\end{table*}

\textbf{Results on different architectures of \boldmath{$T$}}. 
Unlike BiDO, TL-DMI does not require an intensive grid search for hyperparameter selection for a specific architecture. Therefore, TL-DMI offers high practicality and is readily applicable to a range of architectures, whereas existing state-of-the-art MI defenses lack this advantage
\citep{peng2022bilateral}. We conduct evaluations on a range of architectures, including residual-based networks such as ResNet-18/50/101, ResNeSt-101, IR152, as well as the more recent MaxViT architecture \citep{tu2022maxvit}. Across all these experiments in Tab.~\ref{tab:SOTA_highres} and Tab.~\ref{tab:comparison_SOTA}, TL-DMI consistently demonstrate superior performance, highlighting its effectiveness and robustness across various architectures.

\textbf{Result on different \boldmath{${\mathcal{D}_{priv}}$}}}. 
% While the SOTA MI defense \citep{peng2022bilateral} primarily concentrates on the facial dataset CelebA as $\mathcal{D}_{priv}$, 
Regarding private dataset 
$\mathcal{D}_{priv}$, 
in addition to 
CelebA, which is standard for MI research, and other  large-scale facial datasets including Facescrub \citep{ng2014data} and VGGFace2 \citep{cao2018vggface2}, our experiments go beyond these datasets by studying the animal domain, i.e., Stanford Dogs dataset 
\citep{dataset2011novel}. The result is illustrated in Tab.~\ref{tab:SOTA_highres}. Via our comprehensive evaluation, we find that our approach consistently demonstrates its efficacy across various datasets, regardless multiple factors such as the number of training/attack classes or the specific domain under consideration. This versatility highlights the robustness and adaptability of our defense TL-DMI across a wide range of scenarios.

Overall, all these extensive results consistently support that our method is effective in defending against advanced MI attacks.
Our approach is simple and can be easily applied, 
with minimal changes to the original training of target classifier $T$.
 {\bf Additional results and analysis are included in the Supp.}

\section{Conclusion}

In this paper, we propose  a simple and highly effective \underline{T}ransfer \underline{L}earning-based \underline{D}efense against \underline{M}odel \underline{I}nversion (TL-DMI).
Our method is a major departure from existing MI defense based on dependency minimization regularization. Our main idea is to leverage TL to limit the number of layers encoding private data information, thereby degrading the performance of MI attacks. 
To justify our method, we conduct the first study to analyze layer importance for MI task via Fisher Information. Our analysis results suggest that the first few layers are important for MI,  justifying our design to prevent private information encoded in the first few layers.
Our defense TL-DMI is remarkably simple to implement.
Through extensive experiments, we demonstrate SOTA  effectiveness of TL-DMI across 20 MI setups spanning 9 architectures, 4 private datasets $D_{priv}$, and 7 MI attacks.

 {\bf Limitation.} 
Following other MI attack and defense research \cite{zhang2020secret,chen2021knowledge,nguyen_2023_CVPR,wang2021improving,peng2022bilateral}, our current focus is on classification. However, our future work will extend to studying MI attacks and defenses for other machine learning tasks, such as object detection. 

% \vspace{-0.2cm}

 {\bf Ethical consideration.}
Our research on improving MI robustness addresses a significant ethical concern in modern data-driven machine learning: data privacy. Our study is based  on publicly available standard data and does not involve the collection of sensitive information.

\noindent {\bf Acknowledgement.} This research is supported by the National Research Foundation, Singapore under its AI Singapore Programmes (AISG Award No.: AISG2-TC-2022-007); The Agency for Science, Technology and Research (A*STAR) under its MTC Programmatic Funds (Grant No. M23L7b0021). This material is based on the research/work support in part by the Changi General Hospital and Singapore University of Technology and Design, under the HealthTech Innovation Fund (HTIF Award No. CGH-SUTD-2021-004).

{
    \small
    \bibliographystyle{ieeenat_fullname}
    \bibliography{main}
}


% \newpage

\section*{Supplementary Materials}

\section{Additional Results} \label{Addtional results}

\subsection{Additional result on BREPMI}\label{BREPMI}

\begin{table}[ht]
\centering
\centering
\tiny
% \vspace{-0.3cm}
\begin{adjustbox}{width=0.5\textwidth,center}
\begin{tabular}{ccccc}
\hline
\T\B Defense & Acc $\Uparrow$ & AttAcc $\Downarrow$ &  $\Delta \Uparrow$ & KNN $\Uparrow$    \\ \hline
No Def. \T\B & 89.00          & 69.67          & -                                                                & 1337.01 \\ \cline{1-5} 
BiDO \T\B   & 80.35  & 39.73  & 3.46                                                             & 1534.48 \\ \cline{1-5} 
TL-DMI  \T\B  & 83.41  & 42.00 & \textbf{4.90 }                                                            & 1517.38 \\ \hline
\end{tabular}
\label{tab:SOTA_lowres}
\end{adjustbox}
\caption{Empirical results for BREPMI \cite{kahla2022label}. Following the exact experimental setups from BREPMI, $\mathcal{D}_{priv}$ = CelebA, $\mathcal{D}_{pub}$ = CelebA, evaluation model = FaceNet, and target classifier $T$ = VGG16, there are a total of 300 attacked classes. Our proposed TL-DMI achieves better {MI robustness, which is quantified by \bf MI robustness is quantified by the $\Delta$, the ratio of drop in attack accuracy to drop in natural accuracy}}
% \vspace{-0.4cm}
\end{table}

\subsection{Additional Empirical Validation on GMI} \label{Additional Empirical Validation on GMI}

\setlength{\columnsep}{7pt}
\begin{table}
% \vspace{-0.5cm}
% \vspace{-0.8cm}
\begin{adjustbox}{width=0.4\textwidth,center}
\begin{tabular}{cccc}
\hline
   \textbf{Defense}     & \textbf{Acc} $\Uparrow$ & \textbf{AttAcc} $\Downarrow$ & $\mathbf{\Delta}$ $\Uparrow$ \\ \hline
No Def. & 90.55                   & 83.87                        & -                            \\
TL-DMI    & 85.60                   & 19.25                        & \textbf{13.05}               \\ \hline
\end{tabular}
\end{adjustbox}
% \vspace{-0.15cm}
\caption{We follow the MI setup from MIRROR, where $T$ = ResNet-34, $\mathcal{D}_{priv}$ = Stanford Cars, $\mathcal{D}_{pub}$ = LSUN Cars, $\mathcal{D}_{pretrain}$ = ImageNet1K.}
\label{tab:car}
% \vspace{-0.4cm}
\end{table}

 \begin{figure}[t]
 \centering
 \begin{adjustbox}{width=0.5\textwidth,center}
 \includegraphics[width=0.5\textwidth]{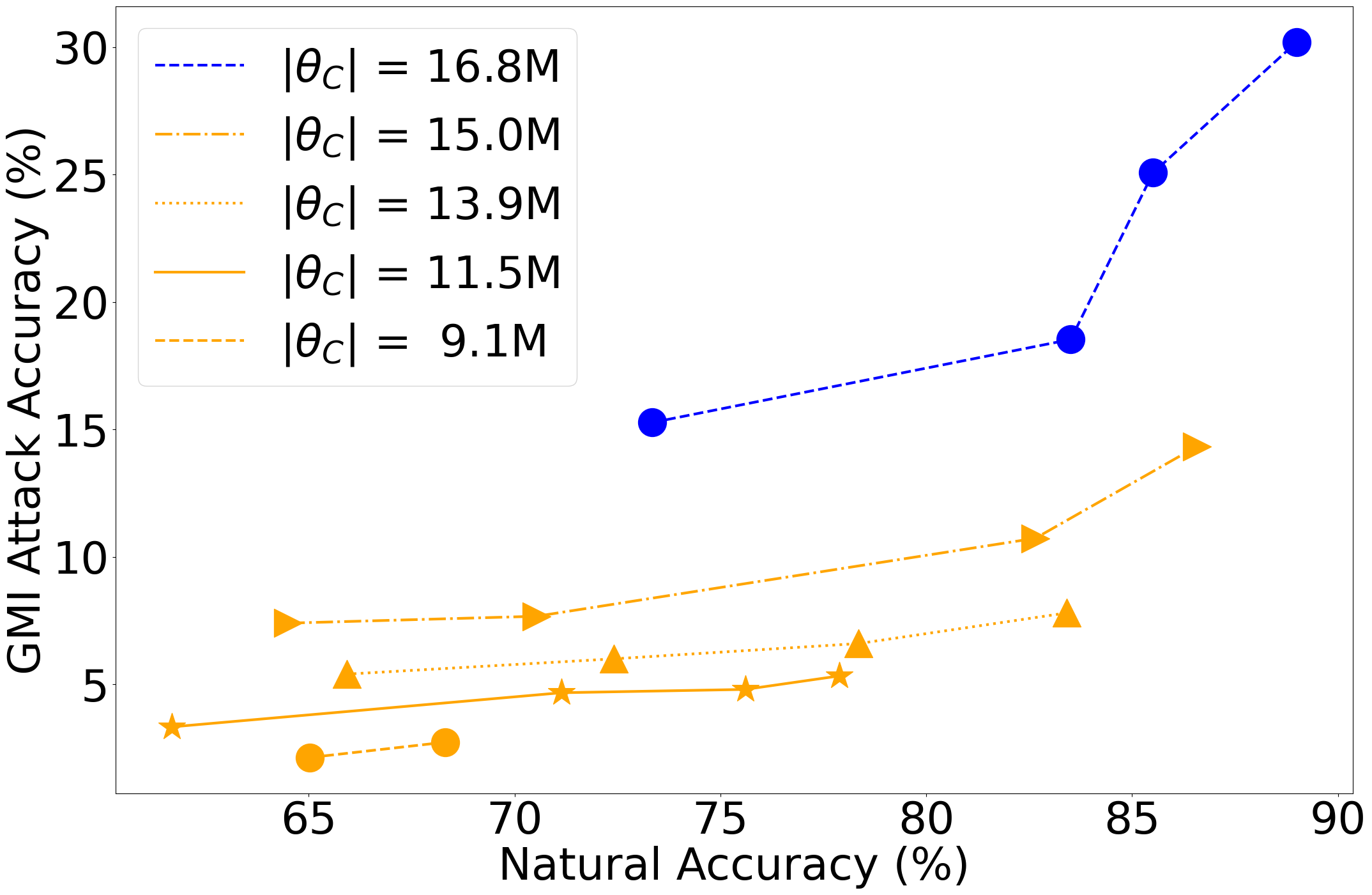}
 \end{adjustbox}
 \caption{Empirical Validation on VGG16 with GMI. Each line represents one training setup for $T$ with a different $|\theta_C|$ updated on $\mathcal{D}_{priv}$. {\color{blue} Note that number of parameters for the entire target model  $|\theta_T| = 16.8$M for this MI setup}. To separate the influence of  natural accuracy on MI attack accuracy, we perform GMI attacks on different checkpoints for each training setup, varying a wide range of natural accuracy. This is presented by multiple data points on each line.  For a given natural accuracy, it can be clearly observed that attack accuracy can be reduced by decreasing $|\theta_C|$, i.e., decreasing parameters updated on $\mathcal{D}_{priv}$.}
 \label{fig:GMI_Empirical}
 % \vspace{-0.5cm}
 \end{figure}

Beside the empirical validation on VGG16 with KEDMI and ResNet-18 with PPA presented in the main manuscript. We also provide additional empirical validation on VGG16 with GMI in Fig.~\ref{fig:GMI_Empirical}. The observation is consistent with the results in the main manuscript.

\subsection{Additional result on LOMMA}\label{Additional LOMMA}

Due to the remarkably simple implementation of our proposed TL-DMI, we expand the MI robustness evaluation to LOMMA \cite{nguyen_2023_CVPR}, is the SOTA MI attacks. Note that this MI attack has not been included yet in SOTA MI defense BiDO. The results in Tab.~\ref{tab:LOMMA_Extensive_setup} shown that TL-DMI is able to defense against SOTA MI attack LOMMA. For a fair comparison, we strictly follow LOMMA for the MI setups.

\subsection{Additional result on Stanford Cars dataset}\label{Additional Stanford Cars}

We further show the effectiveness of our proposed TL-DMI on Dogs breeds classification (see Tab. 4) and additional Cars Classification in Tab.~\ref{tab:car}.
The results further show the effectiveness of TL-DMI.

\setlength{\columnsep}{7pt}
\begin{table}
% \addtolength{\tabcolsep}{-4pt}  
\setlength\tabcolsep{2pt}
% \setlength\extrarowheight{2pt}
% \vspace{-0.65cm}
\begin{adjustbox}{width=0.45\textwidth,center}
\begin{tabular}{ccccc}
\hline
\textbf{Attack} & \textbf{Defense} & \textbf{Acc} $\Uparrow$ & \textbf{AttAcc} $\Downarrow$ & \boldmath{$\Delta$} $\Uparrow$         \\ \hline
\multirow{5}{*}{PPA}   & No Def.          & 94.86        & 82.83 $\pm$ 0.17    & -                     \\
                       & MID (0.05)       & 90.85        & 52.09 $\pm$ 0.45    & 7.67 $\pm$ 0.09           \\
                       & MID (0.02)       & 91.54        & 61.67 $\pm$ 0.33    & 5.77 $\pm$ 0.06           \\
                       & MID (0.01)       & 92.70        & 75.84 $\pm$ 0.60    & 2.11 $\pm$ 0.15           \\
                       & TL-DMI             & 90.10        & 31.70 $\pm$ 0.17    & \textbf{10.74 $\pm$ 0.05} \\ \hline
\multirow{5}{*}{LOMMA} & No Def.          & 89.00        & 93.68 $\pm$ 1.94    & -                     \\
                       & MID (0.002)      & 78.06        & 76.51 $\pm$ 0.27    & 1.57 $\pm$ 0.20           \\
                       & MID (0.003)      & 75.83        & 78.73 $\pm$ 1.88    & 0.28 $\pm$ 0.29           \\
                       & MID (0.004)      & 72.87        & 76.14 $\pm$ 0.90    & 1.09 $\pm$ 0.10           \\
                       & TL-DMI             & 83.41        & 72.47 $\pm$ 2.85    & \textbf{3.79 $\pm$ 0.19}  \\ \hline
\end{tabular}
\end{adjustbox}
% \vspace{-0.1cm}
\caption{Varying MID hyperparameters, we conduct three comparisons, reporting mean and standard deviation.}
% {\color{purple} \textbf{Man:bold the best delta as usual. Show std dev also for GMI/KEDMI.}}
% \vspace{-0.7cm}
% \hspace{-0.5cm}
\label{tab:MID_hyper}
\end{table}

\subsection{Additional results on other MI setups}\label{Additional MI setup}

We further provide 3 more setups with PPA in Tab.~\ref{tab:other_setups} in this rebuttal. All results consistently support outstanding defense trade-off with TL-DMI

\setlength{\columnsep}{7pt}
\begin{table}
% \addtolength{\tabcolsep}{-4pt}  
\setlength\tabcolsep{2pt}
% \setlength\extrarowheight{2pt}
% \hspace{2cm}
% \vspace{-0.5cm}
% \vspace{-0.85cm}
\begin{adjustbox}{width=0.45\textwidth,center}
\begin{tabular}{cccccc}
\hline
\textbf{Attack} & $\mathbf{T}$            & \textbf{Defense}         & \textbf{Acc}     $\Uparrow$         & \textbf{AttAcc}  $\Downarrow$    &    $\mathbf{\Delta}$ $\Uparrow$                   \\ \hline
\multirow{6}{*}{PPA}   & \multirow{2}{*}{ResNet-18}  & No Def.                  & 94.22                     & 47.41 $\pm$ 0.18         & -                     \\
                       &                             & TL-DMI                     & 91.12                     & 5.19 $\pm$ 0.23          & \textbf{13.62 $\pm$ 0.04} \\ \cline{2-6} 
                       & \multirow{2}{*}{ResNet-101} & No Def.                  & 96.57                     & 38.99 $\pm$ 0.19         & -                     \\
                       &                             & TL-DMI                     & 93.01                     & 6.66 $\pm$ 0.12          & \textbf{9.08 $\pm$ 0.03}  \\ \cline{2-6} 
                       & \multirow{2}{*}{MaxViT}     & No Def.                  & 96.57                     & 31.79 $\pm$ 0.20         & -                     \\
                       &                             & \multicolumn{1}{c}{TL-DMI} & \multicolumn{1}{c}{93.01} & \multicolumn{1}{c}{4.11 $\pm$ 0.16} & \multicolumn{1}{c}{\textbf{7.78 $\pm$ 0.03}}  \\ \hline
\end{tabular}
\end{adjustbox}
% \vspace{-0.15cm}
\caption{Additional MI setups for PPA, where $\mathcal{D}_{priv}$ = FaceScrub, $\mathcal{D}_{pub}$ = Metfaces, $\mathcal{D}_{pretrain}$ = ImageNet1K. 
%We present mean and standard deviation of three runnings.
}
% \vspace{-0.5cm}
\label{tab:other_setups}
\end{table}% \vspace{-0.12cm}

\subsection{Comparison with SOTA MI Defense}\label{Comaprison with SOTA}

We provide a comprehensive comparisons between our proposed TL-DMI and BiDO and MID under {\bf 6 attacks}: VMI, LOMMA, PPA, KEDMI, GMI, and BREPMI. To avoid the effect of randomness in our comparison, we calculate $\Delta$ under 3 attacks of different random seeds. We summarize the comparison in Tab.~\ref{tab:SOTA_compare}. \textbf{All results consistently support that TL-DMI outperforms BiDO and MID}

For BiDO reproducibility, we follow the exact hyperparemeters from their work. Note that BiDO is the best defense by far, but it requires extensive grid-search for hyperparameters. For MID reproducibility, we adopt their implementation and hyperparameters. Furthermore, we provide the results for MID with different hyperparameter choices in Tab.~\ref{tab:MID_hyper}.

\begin{table}[]
\begin{adjustbox}{width=0.45\textwidth,center}
\begin{tabular}{cccccc}
\hline
\textbf{Attack}   & $\mathbf{T}$                  & \textbf{Defense} & \textbf{Acc}   $\Uparrow$      & \textbf{AttAcc} $\Downarrow$     & \boldmath{$\Delta$} $\Uparrow$           \\ \hline
\multirow{4}{*}{LOMMA}   & \multirow{4}{*}{VGG-16}     & No Def.          & 89.00                & 93.68 $\pm$ 1.94         & -                    \\
                         &                             & MID              & 78.06                & 76.51 $\pm$ 0.27         & 1.57 $\pm$ 0.20          \\
                         &                             & BiDO             & 80.35                & 66.22 $\pm$ 3.74         & 3.17 $\pm$ 0.29          \\
                         &                             & TL-DMI             & 83.41                & 72.47 $\pm$ 2.85         & \textbf{3.79 $\pm$ 0.19}          \\ \hline
\multirow{8}{*}{PPA}     & \multirow{4}{*}{ResNet-18}  & No Def.          & 94.22                & 90.08 $\pm$ 1.40         & -                    \\
                         &                             & MID              & \multicolumn{1}{c}{88.27} & \multicolumn{1}{c}{48.81 $\pm$ 0.28} & \multicolumn{1}{c}{6.94 $\pm$ 0.22} \\
                         &                             & BiDO             & 91.33                & 76.65 $\pm$ 0.09         & 4.65 $\pm$ 0.46          \\
                         &                             & TL-DMI             & 91.12                & 21.32 $\pm$ 0.90         & \textbf{22.18 $\pm$ 0.74}         \\ \cline{2-6} 
                         & \multirow{4}{*}{ResNet-101} & No Def.          & 94.86                & 82.83 $\pm$ 0.17         & -                    \\
                         &                             & MID              & 90.85                & 52.09 $\pm$ 0.45         & 7.67 $\pm$ 0.09          \\
                         &                             & BiDO             & 90.32                & 67.43 $\pm$ 0.36         & 3.39 $\pm$ 0.09          \\
                         &                             & TL-DMI             & 90.10                & 31.70 $\pm$ 0.17         & \textbf{10.74 $\pm$ 0.05}         \\ \hline
\multirow{4}{*}{KEDMI}   & \multirow{4}{*}{VGG-16}     & No Def.          & 89.00                & 87.71 $\pm$ 2.73         & -                    \\
                         &                             & MID              & 78.06                & 66.64 $\pm$ 0.78         & 1.93 $\pm$ 0.30          \\
                         &                             & BiDO             & 80.35                & 39.77 $\pm$ 5.60         & 5.54 $\pm$ 0.33          \\
                         &                             & TL-DMI             & 83.41                & 51.64 $\pm$ 1.97         & \textbf{6.45 $\pm$ 0.59}          \\ \hline
\multirow{4}{*}{BREP-MI} & \multirow{4}{*}{VGG-16}     & No Def.          & 89.00                & 70.56 $\pm$ 1.84         & -                    \\
                         &                             & MID              & \multicolumn{1}{l}{78.06} & \multicolumn{1}{c}{16.47 $\pm$ 1.07} & \multicolumn{1}{c}{4.91 $\pm$ 0.26} \\
                         &                             & BiDO             & 80.35                & 39.35 $\pm$ 0.90         & 3.61 $\pm$ 0.16          \\
                         &                             & TL-DMI             & 83.41                & 41.22 $\pm$ 0.69         & \textbf{5.24 $\pm$ 0.41}          \\ \hline
\multirow{4}{*}{VMI}     & \multirow{4}{*}{ResNet-34}  & No Def.          & 69.27                & 39.40                & -                    \\
                         &                             & MID              & 52.52                & 29.05                & 0.62                 \\
                         &                             & BiDO             & 61.14                & 30.25                & 1.13                 \\
                         &                             & TL-DMI             & 62.20                & 21.73 $\pm$ 2.08         & \textbf{2.5 $\pm$ 0.30}           \\ \hline
\multirow{4}{*}{GMI}     & \multirow{4}{*}{VGG-16}     & No Def.          & 89.00                & 31.25 $\pm$ 1.04         & -                    \\
                         &                             & MID              & 78.06                & 28.78 $\pm$ 1.24         & 0.25 $\pm$ 0.16          \\
                         &                             & BiDO             & 80.35                & 6.31 $\pm$ 0.54          & 2.88 $\pm$ 0.15          \\
                         &                             & TL-DMI             & 83.41                & 8.47 $\pm$ 0.58          & \textbf{4.08 $\pm$ 0.11}          \\ \hline
\end{tabular}
\end{adjustbox}
% \vspace{-0.15cm}
\caption{We re-run three comparisons, presenting mean and standard deviation. Following VMI setup from BiDO, we encounter code reproducibility issues, and we take 
the best result reported in BiDO paper.}
% {\color{purple} \textbf{Man:bold the best delta as usual. Show std dev also for GMI/KEDMI.}}
% \vspace{-0.5cm}
% \hspace{-0.5cm}
\label{tab:SOTA_compare}
\end{table}

% \usepackage{multirow}
\begin{table*}[]
\setlength{\tabcolsep}{2pt}
% \vspace{-0.2cm}
% \vspace{-0.3cm}
\begin{adjustbox}{width=2.0\columnwidth,center}
\begin{tabular}{>{\centering\arraybackslash}p{1.7cm}ccccp{1.4cm}p{1.6cm}p{1.0cm}>{\centering\arraybackslash}p{2.1cm}>{\centering\arraybackslash}p{2.1cm}p{0.9cm}}
\hline
\textbf{Attack Method}     & \boldmath{$\mathcal{D}_{priv}$}     & \boldmath{$\mathcal{D}_{pub}$}      & \boldmath{$\mathcal{D}_{pretrain}$}         & \boldmath{$T$}  \T\B       & \textbf{Defense Method} & \textbf{\boldmath{$|\theta_C| \slash |\theta_T|$}} \T\B & \textbf{Acc $\Uparrow$} & \textbf{Top1-AttAcc $\Downarrow$} & \textbf{Top5-AttAcc $\Downarrow$} & \textbf{KNN Dist} $\Uparrow$ \\ \hline

\multirow{14}{*}{LOMMA-K} & \multirow{6}{*}{CelebA} & \multirow{6}{*}{CelebA} & \multirow{2}{*}{ImageNet1K}  & \multirow{2}{*}{VGG16}   & No Def. \T\B         & 16.8/16.8            & 89.00           &   95.67 $\pm$ 0.91        &  96.68  $\pm$ 0.01         &   1158     \\

             &             &             &                &              & TL-DMI          & 13.9/16.8            & 83.41           &    \textbf{75.67 $\pm$ 1.83}   &   \textbf{91.68  $\pm$ 0.01}    & \textbf{1304}   \\ \cline{4-11} 
            
            &             &             & \multirow{4}{*}{MS-CelebA-1M} & \multirow{2}{*}{IR152}   & No Def. \T\B         & 62.6/62.6            & 93.52           &    96.40 $\pm$ 0.51       &   99.67 $\pm$ 0.15        &   1038     \\
            
            &             &             &                &              & TL-DMI          & 17.8/62.6            & 86.70           &  \textbf{77.73 $\pm$ 1.57}     & \textbf{94.67 $\pm$ 0.66}      & \textbf{1305}   \\ \cline{5-11} 
            
            &             &             &                & \multirow{2}{*}{FaceNet64} & No Def. \T\B         &
            35.4/35.4            & 88.50           &    89.33  $\pm$ 1.19      &     98.67 $\pm$ 0.15      &  1226      \\
            
             &             &             &                &              & TL-DMI          & 34.4/35.4               
            &    83.41            &    \textbf{79.60 $\pm$ 1.78}           &     \textbf{97.00 $\pm$ 0.61}           &  \textbf{1345}      \\ \cline{2-11} 
            
            & \multirow{6}{*}{CelebA} & \multirow{6}{*}{FFHQ}  & \multirow{2}{*}{ImageNet1K}  & \multirow{2}{*}{VGG16}   & No Def. \T\B         & 16.8/16.8            & 89.00           &   58.60 $\pm$ 1.67    &  86.00 $\pm$ 1.14       &  1390      \\
            
            &             &             &                &              & TL-DMI          & 13.9/16.8            & 83.41           &   \textbf{36.00 $\pm$ 1.28}    & \textbf{65.00 $\pm$ 1.95}  & \textbf{1550}   \\ \cline{4-11} 
            
            &             &             & \multirow{4}{*}{MS-CelebA-1M} & \multirow{2}{*}{IR152}   & No Def. \T\B         & 62.6/62.6            & 93.52           &      73.47 $\pm$ 1.30  &  90.00 $\pm$ 0.85     &  1290      \\
            
            &             &             &                &              & TL-DMI          & 17.8/62.6            & 86.70           &   \textbf{45.27 $\pm$ 1.98}    & \textbf{74.33 $\pm$ 1.25}      &  \textbf{1474}  \\ \cline{5-11} 
            
            &             &             &                & \multirow{2}{*}{FaceNet64} & No Def. \T\B         & 35.4/35.4            & 88.50           &   70.27 $\pm$ 1.63       &  90.33 $\pm$ 0.72        &   1391     \\
            
     &             &             &                &              & TL-DMI          & 34.4/35.4                &    83.41         &    \textbf{19.53 $\pm$ 1.19}            &    \textbf{41.33 $\pm$ 1.34}            &   \textbf{1759}      \\ \hline

\multirow{14}{*}{LOMMA-G} & \multirow{6}{*}{CelebA} & \multirow{6}{*}{CelebA} & \multirow{2}{*}{ImageNet1K}  & \multirow{2}{*}{VGG16}   & No Def. \T\B         & 16.8/16.8            & 89.00           &   56.00 $\pm$ 3.65        &      79.00 $\pm$ 3.84     &   1454     \\

             &             &             &                &              & TL-DMI          & 13.9/16.8            & 83.41           &   \textbf{22.00 $\pm$ 4.77}    &  \textbf{45.33 $\pm$ 9.08}     &  \textbf{1709}  \\ \cline{4-11} 
            
            &             &             & \multirow{4}{*}{MS-CelebA-1M} & \multirow{2}{*}{IR152}   & No Def. \T\B         & 62.6/62.6            & 93.52           &    64.67 $\pm$ 5.54       &      86.00 $\pm$ 5.09     &   1401     \\
            
            &             &             &                &              & TL-DMI          & 17.8/62.6            & 86.70           &   \textbf{41.87 $\pm$ 5.37}    &   \textbf{70.67 $\pm$ 5.97}    &  \textbf{1551}  \\ \cline{5-11} 
            
            &             &             &                & \multirow{2}{*}{FaceNet64} & No Def. \T\B         & 35.4/35.4            & 88.50           &   60.00  $\pm$ 5.90   &     80.00 $\pm$ 3.81     &  1501      \\
            
     &             &             &                &              & TL-DMI          & 34.4/35.4                &    83.41         &   \textbf{43.67 $\pm$ 5.60}             &       \textbf{65.00 $\pm$ 6.82}         &   \textbf{1616}      \\ \cline{2-11} 
            
            & \multirow{6}{*}{CelebA} & \multirow{6}{*}{FFHQ}  & \multirow{2}{*}{ImageNet1K}  & \multirow{2}{*}{VGG16}   & No Def. \T\B         & 16.8/16.8            & 89.00           &   27.00 $\pm$ 6.10     &   52.33 $\pm$ 5.82        &  1642      \\
            
            &             &             &                &              & TL-DMI          & 13.9/16.8            & 83.41           &   \textbf{8.87 $\pm$ 3.12}    & \textbf{24.00 $\pm$ 5.50}   &  \textbf{1829}  \\ \cline{4-11} 
            
            &             &             & \multirow{4}{*}{MS-CelebA-1M} & \multirow{2}{*}{IR152}   & No Def. \T\B         & 62.6/62.6            & 93.52           &    45.20 $\pm$ 4.30    &  70.67 $\pm$ 4.58     &  1503      \\
            
            &             &             &                &              & TL-DMI          & 17.8/62.6            & 86.70           &  \textbf{22.87 $\pm$ 5.05}    &  \textbf{43.67 $\pm$ 7.46}     & \textbf{1650}   \\ \cline{5-11} 
            
            &             &             &                & \multirow{2}{*}{FaceNet64} & No Def. \T\B         & 35.4/35.4            & 88.50           &     30.60 $\pm$ 5.21     &   62.00 $\pm$ 5.69        &  1625      \\
            
            &             &             &                &              & TL-DMI          & 34.4/35.4                &    83.41            &   \textbf{ 9.33 $\pm$ 4.55}      &   \textbf{24.33 $\pm$ 4.55}             &    \textbf{1909}    \\ \hline
            
\end{tabular}
\end{adjustbox}
\caption{Our extended MI robustness evaluation on SOTA MI attack LOMMA \cite{nguyen_2023_CVPR}. The results of AttAcc and Acc are given in \%. We reports the MI defense results against different LOMMA attack setups including LOMMA+KEDMI (LOMMA-K) and LOMMA+GMI (LOMMA-G) with the varying in different public datasets $\mathcal{D}_{pub}$ (CelebA and FFHQ), and pre-trained datasets $\mathcal{D}_{pretrain}$ (Imagenet1K and MS-CelebA-1M). }
\label{tab:LOMMA_Extensive_setup}
% \vspace{-0.5cm}
\end{table*}
\section{Additional Analysis} \label{Pre-trained datasets}

\subsection{The effect of pretrain dataset to MI robustness}

\begin{figure}[h]
  \centering
  \begin{adjustbox}{width=0.5\textwidth,center}
  \includegraphics[width=0.5\textwidth]{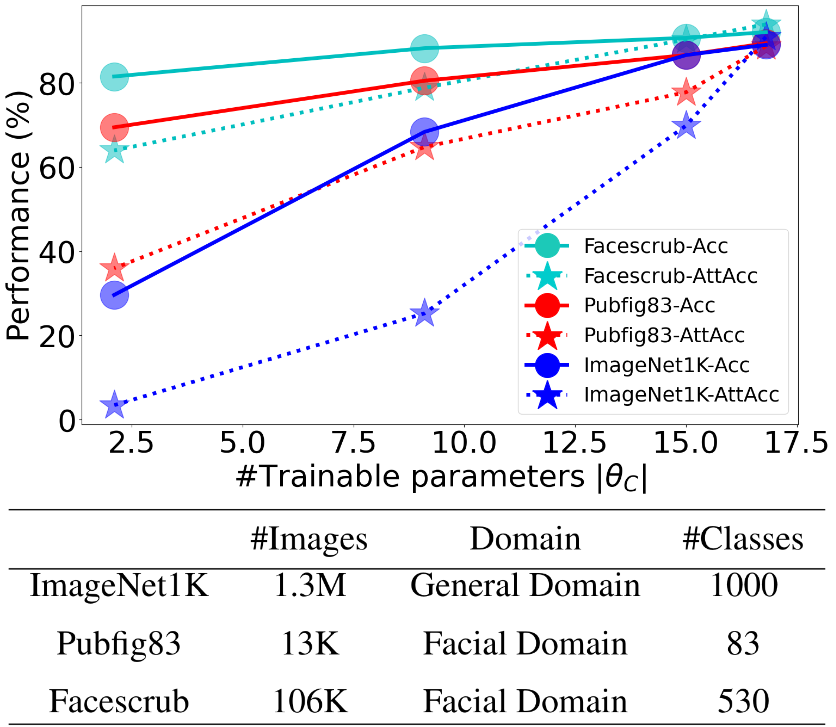}
  \end{adjustbox}
  \caption{The effect of different $\mathcal{D}_{pretrain}$, i.e., ImageNet1K, Pubfig83, and Facescrub. We use $T$ = VGG16, $\mathcal{D}_{priv}$ = CelebA. The results suggest that the less similarity between pretrain and private dataset domains can improve defense effectiveness.}
  \label{fig:Dataset}
  % \vspace{-0.5cm}
\end{figure}

In these above sections, we use a consistent and standard pre-trained dataset to ensure fair comparison with other methods in the literature. Since the pre-trained backbone can be produced with different datasets in practice, we investigate the impact of different pre-trained datasets on MI robustness in this section. Specifically, we implement the same setup as the KEDMI setup for VGG16, but vary three different pre-trained datasets: ImageNet1K, Facescrub, and Pubfig83. The results are shown in Fig.~\ref{fig:Dataset}.

Updating all parameters $|\theta_C|$ = 16.8M on $\mathcal{D}_{priv}$, yields no significant differences among different $\mathcal{D}_{pretrain}$. This is expected and align with our understanding, where all the parameters in $T$ are exposed to private data during the training of $T$. With fewer trainable parameters on $\mathcal{D}_{priv}$, we notice clearer differences. Overall, pre-training on a closer domain (Pubfig83 and Facescrub) restores natural accuracy much better than pre-training on a general domain (Imagenet1K).

For instance, with $|\theta_C|$ = 2.1M, pre-training on Facescrub and Pubfig83 achieve 81.48\% and 69.41\% accuracy, respectively, compared to 29.59\% in the Imagenet1K setup. 
Nevertheless, pre-training on a closer domain also increases the risk of MI attack. As those frozen parameters during the fine-tuning on $\mathcal{D}_{priv}$ keep the feature representations from $\mathcal{D}_{pretrain}$, thus, the closer the $\mathcal{D}_{pretrain}$, the riskier it is for the model against MI attack. Notably, with $|\theta_C|$ = 15.0M, models pre-training on ImagNet1K and Pubfig83 achieve comparable accuracy. However, using ImageNet1K as $\mathcal{D}_{pretrain}$ renders a more robust model (decreasing MI attack accuracy by 8.06\%) than the setup of Pubfig83. In conclusion, when using our TL-DMI to train a MI robust model, it is critical to choose the $\mathcal{D}_{pretrain}$ for a trade-off between restoring model utility and robustness. Specifically, \textbf{less similarity between pretrain and private dataset domains can improve defense effectiveness.}

\subsection{Layer-wise MI Vulnerability Analysis} \label{layer-wise MI vulnerability analysis}
\begin{figure}
\centering
    % \vspace{-0.3cm}
    \includegraphics[width=0.48\textwidth]{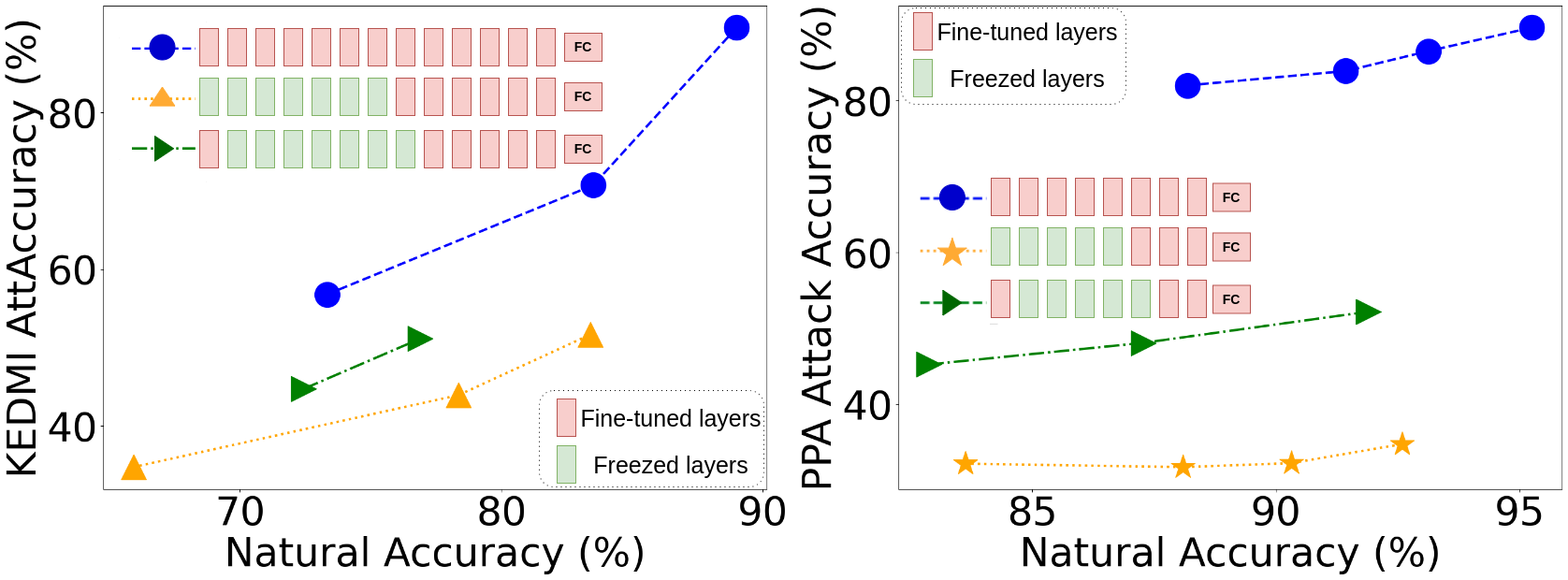}
    % \vspace{-0.1cm}
    \caption{We follow KEDMI-VGG16 and PPA-ResNet-18 setups in Fig. 1-II. Fine-tuning first layers ({\color[HTML]{008118}green line}), rather than middle layers ({\color[HTML]{ffaa44}orange line}), enhances MI attack accuracy, corroborating our analysis: 
    first layers are important for MI. 
    %lowers the natural acc but enhances MI attack accuracy.
    }
    % \vspace{-0.7cm}
    \label{fig:analysis}
\end{figure}

we conduct the following experiments which strongly corroborate our analytical results. Specifically, instead of fine-tuning the middle layers, we fine-tune the first layers, see Fig.~\ref{fig:analysis} in this rebuttal. This single change significantly degrades the defense performance and helps MI attacks, which corroborate our analytical results: first layers are important for MI based on our Fisher Information analysis; therefore, fine-tuning the first layers with private dataset helps MI attacks significantly. 
As another detail to further corroborate our analysis, we remark that first layers have less parameters than middle layers. Yet, MI attacks perform better with fine-tuning private dataset in first layers. This further supports first layers are important for MI. We remark that last layers are critical for classification task, consistent with TL literature. The natural accuracy is much degraded if fine-tuning of last layers is removed.

\subsection{Additional Analysis of Layer Importance} \label{Addtional FI}

\textbf{FI across MI iterations. } MI is a multiple iteration process. The FI for MI in the main manuscript is computed at the last iteration (the iteration that we present the result throughout our submission). Fig.~\ref{fig:FI_iter} also provides the FI across multiple iterations. We observe that after a few iterations, the FI for earlier layers keeps dominant compared to the later layers.

\textbf{Different MI losses. } In the main manuscript, we use $l_2$ distance to compute the MI loss. In addition, we provide FI results using $l_1$  distance and LPIPS \citep{zhang2018unreasonable} to compute the MI loss. The FI results obtained using different MI loss functions are consistent with our main FI observation in the main manuscript.

These additional FI results are consistent with those in our main FI observation in the main manuscript.

\begin{figure*}[h]
  \centering
  \begin{adjustbox}{width=1.0\textwidth,center}
  \includegraphics[width=1.0\textwidth]{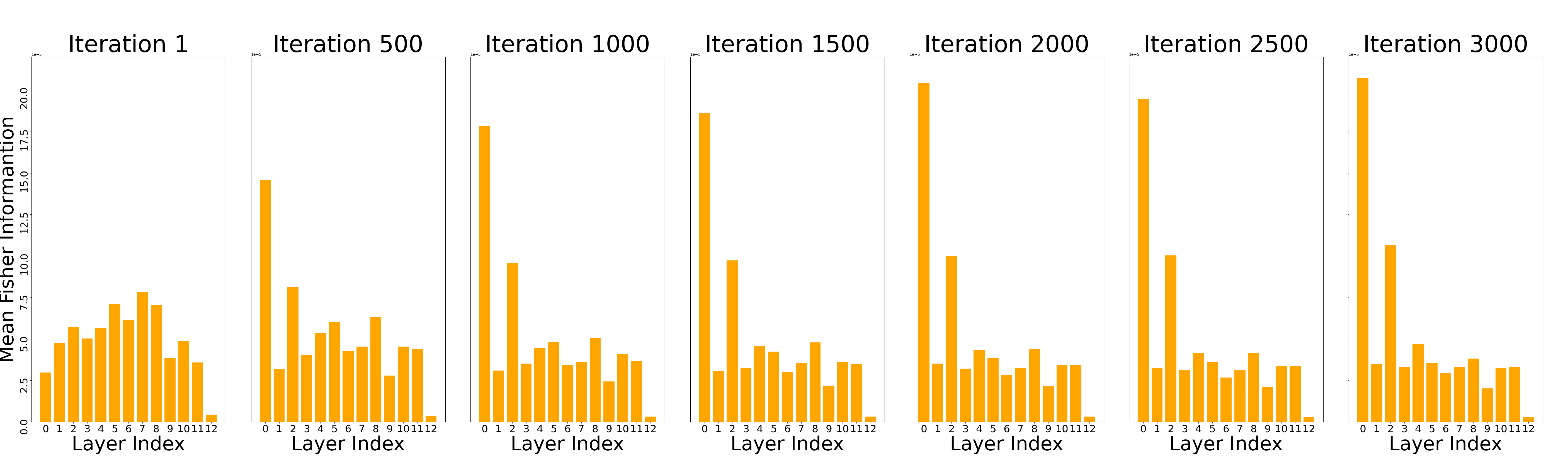}
  \end{adjustbox}
  % \vspace{-0.5cm}
  \caption{FI distributions across layers during all MI steps. We conduct FI analysis on the main setup in Peng et al \cite{peng2022bilateral} where the MI attack is KEDMI \cite{chen2021knowledge}, $T$=VGG16, $\mathcal{D}_{priv}$=CelebA and $\mathcal{D}_{pub}$=CelebA. In the main manuscript, we present the FI analysis at the last MI iteration, i.e., iteration 3000. This figures present a more comprehensive FI analysis across multiple iterations. After first few iterations, we consistently observe that the earlier layers are more important to MI task.}
  \label{fig:FI_iter}
  % \vspace{-0.5cm}
  
\end{figure*}

\begin{figure*}[h]
  \centering
  \begin{adjustbox}{width=1.0\textwidth,center}
  \includegraphics[width=1.0\textwidth]{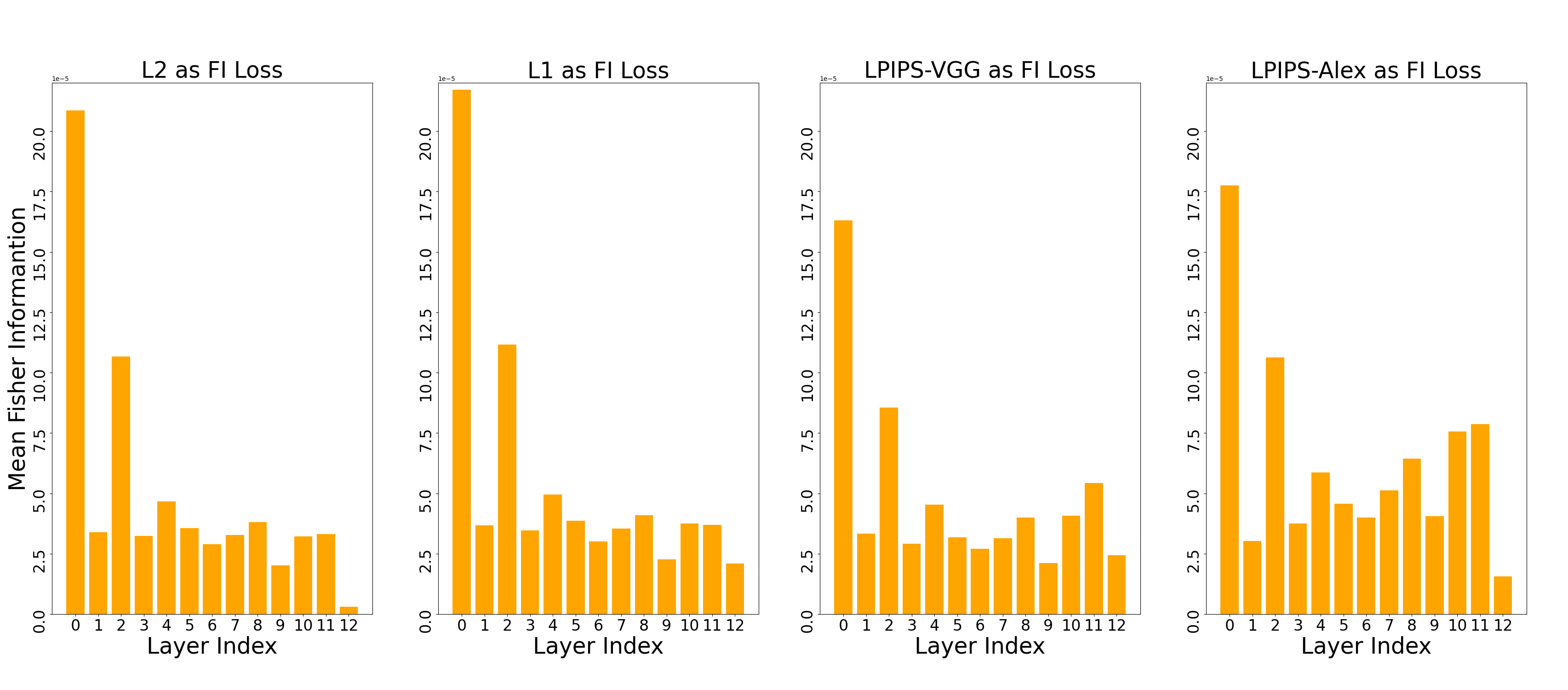}
  \end{adjustbox}
  % \vspace{-0.5cm}
  \caption{FI distributions across layers via different FI losses. We conduct FI analysis on the main setup in Peng et al \cite{peng2022bilateral} where the MI attack is KEDMI \cite{chen2021knowledge}, $T$=VGG16, $\mathcal{D}_{priv}$=CelebA and $\mathcal{D}_{pub}$=CelebA. In the main manuscript, we use $l_2$ distance between reconstructed images and private images as MI loss for the FI analysis. This figure presents the FI analysis through other distances including $l_1$, LPIPS-VGG \cite{zhang2018unreasonable}, LPIPS-ALEX \cite{zhang2018unreasonable}. The results show the consistent observation that the earlier layers of a network are more important to MI attacks compared with later layers.}
  \label{fig:FI_iter}
  % \vspace{-0.5cm}
\end{figure*}

\subsection{MI Robustness via the False Positive Concept}

We provide additional analysis in this Appendix to provide a clear understanding of how our proposed TL-DMI effectively defends against MI attacks, leading to more false positive during MI attacks and decrease in attack accuracy.

As discussed, it has been shown that when a deep neural network-based classifier, denoted as $T = C \circ E$, is pre-trained on a large-scale dataset $\mathcal{D}_{pretrain}$, the features learned in the earlier layers $E$ are transferable to another somewhat related classifier on datasets $\mathcal{D}_{priv}$, enabling the model to maintain its natural accuracy without explicitly updating its parameters on $D_{priv}$ in the earlier layers \citep{yosinski2014transferable}. This transferability of features benefits our proposed TL-DMI through maintaining the model classification performance and natural accuracy.

In contrast, MI attacks require  accurate features to reconstruct the private training dataset $\mathcal{D}_{priv}$. By refraining from updating $E$ on $\mathcal{D}_{priv}$, we limit the leakage of private features into $E$, thereby improving MI robustness. Specifically, recall MI attacks are usually formulated as:

\begin{equation}
\label{eq:white_box_latent1}
 w^* = \arg \min_w (- \log P_T (y|G(w)) + \lambda \mathcal{L}_{prior}(w))
\end{equation}

\begin{figure*}[ht]
\centering
\begin{adjustbox}{width=1.0\textwidth,center}
\includegraphics[width=1.0\textwidth]{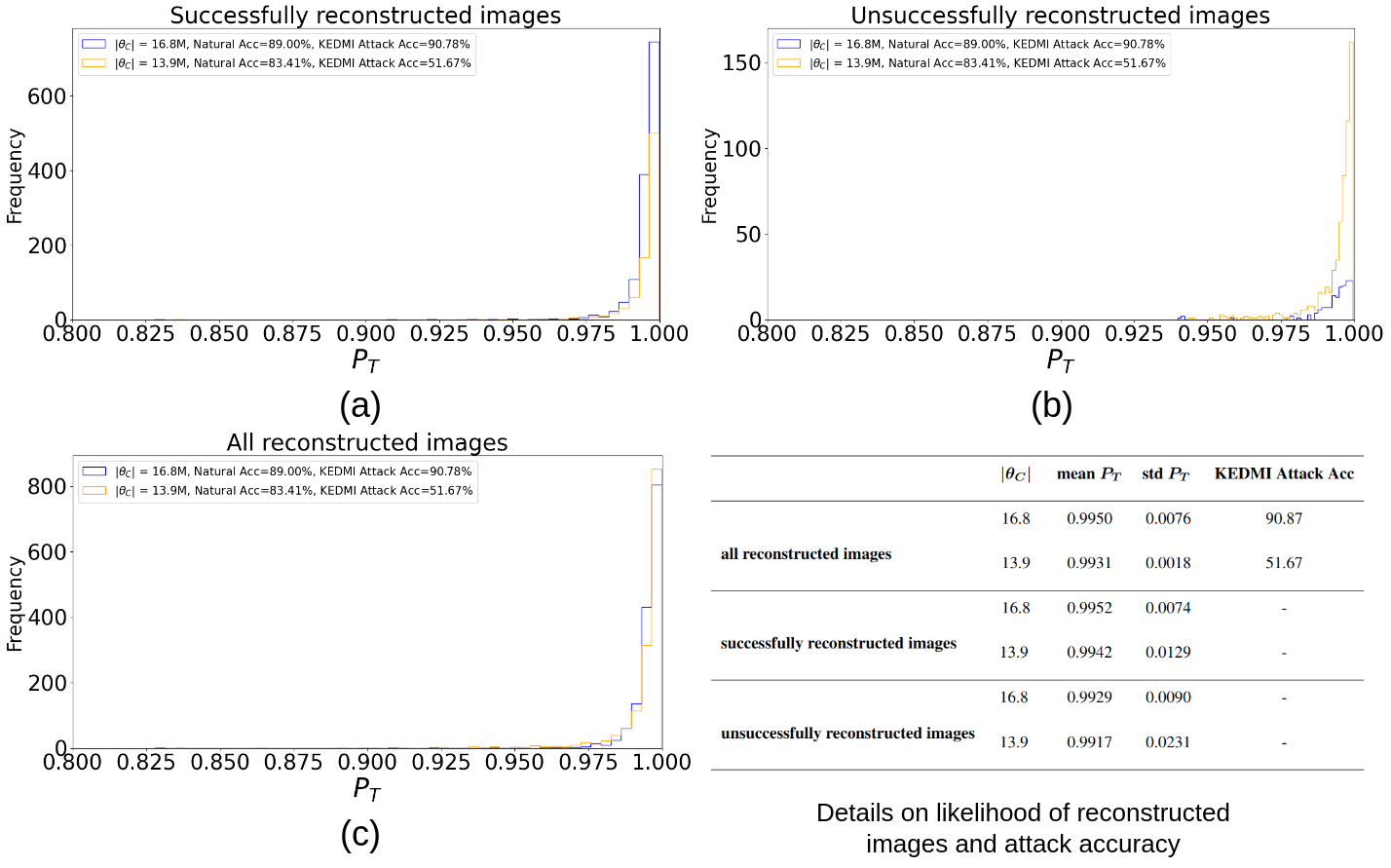}
\end{adjustbox}
% \vspace{-0.5cm}
\caption{Visualization of the distribution of $P_T$ for two models: the {\color{blue} no defense model} (with $|\theta_C| = 16.8M$) and TL-DMI {\color{orange} proposed approach} (with $|\theta_C| = 13.9M$). The visualization is conducted using KEDMI as the attack method, with $\mathcal{D}_{priv}$ = CelebA, $\mathcal{D}_{pub}$ = CelebA, $\mathcal{D}_{pretrain}$ = Imagenet1K, and $T$ = VGG16. We observe that both our proposed TL-DMI model and the model without defense exhibit similar distributions of $P_T$. The values of $P_T$ for both successfully and unsuccessfully reconstructed images are very close to 1 in both cases. However, the attack accuracy shows a significant drop from 90.87 to 51.67 when our proposed TL-DMI is applied.}
% \vspace{-0.4cm}
\label{fig:PT_KEDMI}
\end{figure*}

\begin{figure*}[ht]
\centering
\begin{adjustbox}{width=1.0\textwidth,center}
\includegraphics[width=1.0\textwidth]{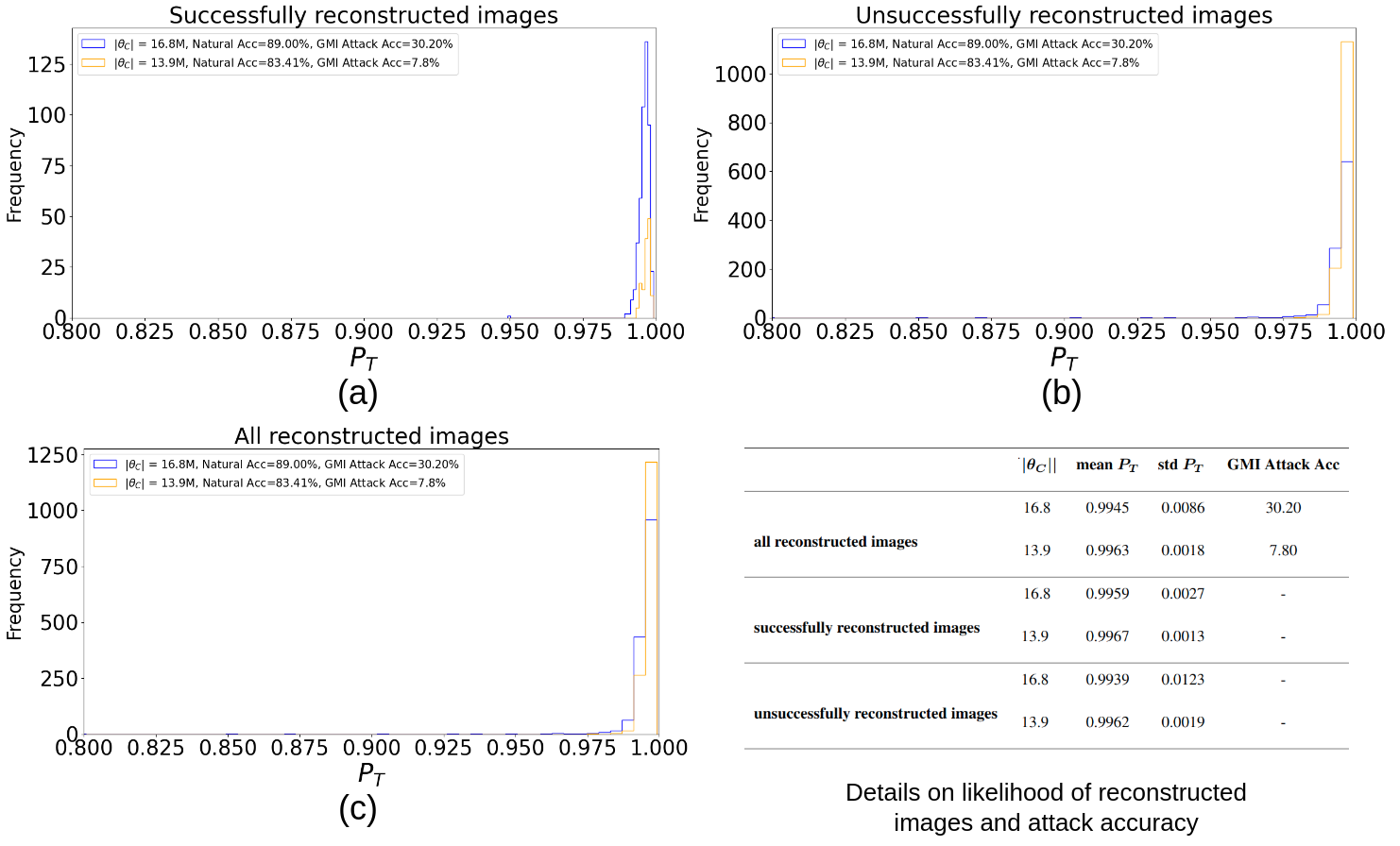}
\end{adjustbox}
% \vspace{-0.5cm}
\caption{Visualization of the distribution of $P_T$ for two models: the {\color{blue} no defense model} (with $|\theta_C| = 16.8M$) and TL-DMI {\color{orange} proposed approach} (with $|\theta_C| = 13.9M$). The visualization is conducted using GMI as the attack method, with $\mathcal{D}_{priv}$ = CelebA, $\mathcal{D}_{pub}$ = CelebA, $\mathcal{D}_{pretrain}$ = Imagenet1K, and $T$ = VGG16. We observe that both our proposed TL-DMI model and the model without defense exhibit similar distributions of $P_T$. The values of $P_T$ for both successfully and unsuccessfully reconstructed images are very close to 1 in both cases. However, the attack accuracy shows a significant drop from 90.87\% to 51.67\% when our proposed TL-DMI is applied.}
% \vspace{-0.4cm}
\label{fig:PT_GMI}
\end{figure*}

Therefore, MI attacks aim to seek $w$ with high likelihood $P_T (y|G(w))$.
We make this key observation to understand how our proposed TL-DMI can degrade MI task: {\em With TL-DMI defense,
while latent variables with high likelihood $P_T (y|G(w))$ can still be identified via
Eq. \ref{eq:white_box_latent1}, 
many $w^*$ are false positives, i.e. $G(w^*)$ do not resemble private samples.
This results in decrease in attack accuracy.}
% the \textit{MI optimizations are solved equally well for both our defense model and the model without defense}. 
This can be observed from the likelihood distributions {\color{blue} $P_{T_{|\theta_C|=16.8M}}$} and {\color{orange} $P_{T_{|\theta_C|=13.9M}}$} for both KEDMI (see Fig.~\ref{fig:PT_KEDMI}) and GMI (see Fig.~\ref{fig:PT_GMI}), which are similar and close to 1. These findings indicate that with TL-DMI,  Eq.~\ref{eq:white_box_latent1} could still perform well to seek 
%the optimization for MI still works effectively in terms of optimizing the 
latent variables $w$ to maximize the likelihood $P_T(y|G(w))$. 
However, although likelihood distributions {\color{blue} $P_{T_{|\theta_C|=16.8M}}$} and {\color{orange} $P_{T_{|\theta_C|=13.9M}}$} 
are similar under attacks, the attack accuracy of model with {\color{orange} $|\theta_C|= 13.9M$} is significantly lower than that with {\color{blue} $|\theta_C|= 16.8M$}. This 
suggests that, due to lack of private data information in $E$ in our proposed TL-DMI model {\color{orange} $|\theta_C|= 13.9M$}, many $w^*$ do not correspond to images resembling private images.

% discrepancy leads to interesting insights regarding the behavior of optimized latent variables $w$.

In the setup where {\color{blue} $|\theta_C|= 16.8M$}, the optimization process causes the latent variables $w$ to converge towards regions that are closer to the private samples. This outcome is expected since the model possesses richer low-level features from the private dataset $D_{priv}$ in both $E$ and $C$. Consequently, we observe more true positives after MI optimization, where the likelihood $P_T(y|G(w))$ is well maximized, and the evaluation model successfully classifies them as label $y$.

In contrast, in the setup where {\color{orange} $|\theta_C|= 13.9M$}, the lack of low-level features from $\mathcal{D}_{priv}$ in $E$ hinders the optimization process. As a result, we observe a higher number of false positives after MI optimization. Although these instances successfully maximize the likelihood $P_T(y|G(w))$, the evaluation model is unable to classify them as label $y$ correctly. Therefore, this behavior indicates a higher level of robustness against the MI attack.

\section{The limitation of Existing MI Defenses} \label{BiDO Limitation}

\textbf{Conflicting objectives between classification and MI defense regularizers:} One limitation of the existing MI defenses \citep{wang2021improving, peng2022bilateral} is the introduction of additional regularizers that conflict with the primary objective of minimizing the classification loss \citep{peng2022bilateral}. This conflict often leads to a significant decrease in the overall model utility.

\textbf{BiDO is sensitive to hyper-parameters}. BiDO \citep{peng2022bilateral}, while attempting to partially recover model utility, suffers from sensitivity to hyper-parameters. Optimizing three objectives simultaneously is a complex task, requiring careful selection of weights to balance the three objective terms. The Tab.~\ref{tab:SOTA_limitation} results in an explicit accuracy drop when adjusting hyper-parameters $\lambda_x$ and $\lambda_y$ even with a small change. The optimized values for $\lambda_x$ and $\lambda_y$ in BiDO are obtained through a grid search \citep{peng2022bilateral}. For example, in the case of BiDO-HSIC, the authors tested values of $\lambda_x \in [0.01, 0.2]$ and $\frac{\lambda_y}{\lambda_x} \in [5, 50]$. Furthermore, BiDO requires an additional parameter, $\sigma$, for applying Gaussian kernels to inputs $x$ and latent representations $z$ in order to utilize COCO \citep{gretton2005kernel} and HSIC \citep{gretton2005measuring} as dependency measurements.

% \usepackage{multirow}
\begin{table*}[t]

\begin{adjustbox}{width=2.00\columnwidth,center}
\begin{tabular}{cccccccccc}
\hline
{\color[HTML]{000000} }                                        & {\color[HTML]{000000} }                                     & \multicolumn{2}{c}{{\color[HTML]{000000} \textbf{First run}}} \T\B       & \multicolumn{2}{c}{\textbf{Second run}}       & \multicolumn{2}{c}{\textbf{Third run}}        & \multicolumn{2}{c}{\textbf{Average}}          \\
\multirow{-2}{*}{{\color[HTML]{000000} \textbf{Architecture}}} & \multirow{-2}{*}{{\color[HTML]{000000} \textbf{MI Attack}}} & {\color[HTML]{000000} \textbf{Natural Acc}} $\Uparrow$   & \textbf{Attack Acc} $\Downarrow$ & \textbf{Natural Acc} $\Uparrow$ \T\B   & \textbf{Attack Acc} $\Downarrow$  & \textbf{Natural Acc}  $\Uparrow$  & \textbf{Attack Acc} $\Downarrow$ & \textbf{Natural Acc} $\Uparrow$   & \textbf{Attack Acc} $\Downarrow$ \\ \hline
{\color[HTML]{000000} }                                        & {\color[HTML]{000000} KEDMI}  \T\B                              & {\color[HTML]{000000} }                        & 51.67 $\pm$ 3.93        &                         & 49.67 $\pm$ 4.86        &                         & 53.60 $\pm$ 4.06        &                         & 51.65 $\pm$ 4.28        \\
\multirow{-2}{*}{{\color[HTML]{000000} \textbf{VGG16}}}        & {\color[HTML]{000000} GMI}    \T\B                              & \multirow{-2}{*}{{\color[HTML]{000000} 83.41}} & 7.80 $\pm$ 3.36         & \multirow{-2}{*}{83.11} & 8.80 $\pm$ 2.28         & \multirow{-2}{*}{83.54} & 8.80 $\pm$ 3.36         & \multirow{-2}{*}{83.35} & 8.47 $\pm$ 3.00         \\ \hline
\textbf{Resnet-34}         \T\B                                    & VMI                                                         & 62.2                                           & 23.70 $\pm$ 21.38       & 62.88                   & 19.55 $\pm$ 12.90       & 63.12                   & 21.95 $\pm$ 12.36       & 62.73                   & 21.73 $\pm$ 15.91       \\ \hline
                                                               & KEDMI    \T\B                                                   &                                                & 64.60 $\pm$ 4.93        &                         & 71.6 $\pm$ 4.85         &                         & 69.33 $\pm$ 5.03        &                         & 68.51 $\pm$ 4.94        \\
\multirow{-2}{*}{\textbf{IR152}}                               & GMI   \T\B                                                      & \multirow{-2}{*}{86.7}                         & 8.93 $\pm$ 3.73         & \multirow{-2}{*}{86.47} & 9.47 $\pm$ 2.57         & \multirow{-2}{*}{86.37} & 9.60 $\pm$ 4.16         & \multirow{-2}{*}{86.51} & 9.33 $\pm$ 3.49         \\ \hline
                                                               & KEDMI     \T\B                                                  &                                                & 73.40 $\pm$ 4.10        &                         & 76.27 $\pm$ 4.09        &                         & 76.20 $\pm$ 3.96        &                         & 75.29 $\pm$ 4.05        \\
\multirow{-2}{*}{\textbf{FaceNet64}}                                      & GMI    \T\B                                                     & \multirow{-2}{*}{83.61}                        & 15.73 $\pm$ 4.58        & \multirow{-2}{*}{83.01} & 15.93 $\pm$ 5.20        & \multirow{-2}{*}{82.71} & 13.6 $\pm$ 3.97         & \multirow{-2}{*}{83.11} & 15.09 $\pm$ 4.58        \\ \hline
\end{tabular}
\end{adjustbox}
\caption{We present the results for running experiments multiples time to show the reproducibility of our proposed TL-DMI. For KEDMI \citep{chen2021knowledge}/GMI \citep{zhang2020secret}, we conduct the attacks with $\mathcal{D}_{priv}$ = CelebA, $\mathcal{D}_{pub}$ = CelebA, $\mathcal{D}_{pretrain}$ = Imagenet1K, and $T$ = VGG16/IR152/FaceNet64. For VMI \citep{wang2021variational}, we conduct the attacks with $\mathcal{D}_{priv}$ = CelebA, $\mathcal{D}_{pub}$ = CelebA, $T$ = Resnet-34, and there is no $\mathcal{D}_{pretrain}$ for this setup.}
\label{tab:error_bar}
\end{table*}

% \usepackage{multirow}
\begin{table*}[ht]

\begin{adjustbox}{width=2.0\columnwidth,center}
\begin{tabular}{ccccccccc}
\hline
\textbf{Architecture} & \textbf{Dataset} & \textbf{Input Resolution} & \textbf{\#Epoch} \T\B & \textbf{Batch size} & \textbf{Learning rate} & \textbf{Optimizer}   & \textbf{Weight Decay} & \textbf{Momentum}    \\ \hline

VGG16  \T\B    & CelebA & 64x64           & 200              & 64                  & 0.02                   & SGD & 0.0001                & 0.9 \\ 

IR152    & CelebA & 64x64       \T\B       & 100              & 64                  & 0.01                  &      SGD                & 0.0001                &       0.9               \\ 

FaceNet64   \T\B    & CelebA & 64x64       & 200              & 8                   & 0.008                  &    SGD                  & 0.0001                &       0.9               \\ 

Resnet-34  \T\B    & CelebA & 64x64        & 200              & 64                  & 0.1                    &     SGD                 & 0.0005                &      0.9                \\ 

Resnet-18 \T\B &   CelebA & 224x224 &  100 & 128 & 0.001 & Adam & - & - \\
MaxViT \T\B & CelebA & 224x224 &   100 & 64 & 0.001 & Adam & - & - \\
ResNeSt-101 \T\B &  Stanford Dogs & 224x224 &  100 & 128 & 0.001 & Adam & - & - \\
Resnet-50 \T\B &  VGGFace2 & 224x224 &  100 & 1024 & 0.001 & Adam & - & - \\
\hline
\end{tabular}
\end{adjustbox}
\label{tab:training_settings}
\caption{Training settings for target classifier $T$. We follow the procedure for training $T$ from previous works \cite{chen2021knowledge,nguyen_2023_CVPR,struppek2022plug}}
\label{tab:training_settings}.
% \vspace{-0.6cm}
\end{table*}

\section{Experiment Setting} \label{detail setup}

\subsection{Detailed MI Setup}

\textbf{Attack Dataset.} Following existing MI works \citep{zhang2020secret, wang2021variational, chen2021knowledge, nguyen_2023_CVPR}, our work forcuses on the study of CelebA \citep{liu2015deep}. Furthermore we demonstrate the efficacy of our proposed TL-DMI on other facial datasets with more attack classes (Facescrub \citep{ng2014data}) or larger scale (VGGFace2 \citep{cao2018vggface2}) and on the animal dataset Stanford Dogs \citep{KhoslaYaoJayadevaprakashFeiFei_FGVC2011}. The details for these datasets used in the experimental setups can be found in Tab.~\ref{tab:MI_Dataset_Setup}.

\textbf{Attack Data Preparation Protocol.} Following previous works \citep{zhang2020secret, chen2021knowledge, wang2021variational, nguyen_2023_CVPR, an2022mirror, struppek2022plug} approaches, we split the dataset into private $\mathcal{D}_{priv}$ and public $\mathcal{D}_{pub}$ subsets with no class intersection. $\mathcal{D}_{priv}$ is used to train the target classifier T, while $\mathcal{D}_{pub}$ is used to extract general features only.

\textbf{Target Classifier \boldmath{$T$}.} We select VGG16 for $T$ for a fair comparison with SOTA MI defense \citep{peng2022bilateral}. As our proposed TL-DMI is architecture-agnostic, we also extend the defense results on more common and recent architectures: i.e., IR152 \citep{he2016deep}, FaceNet64 \citep{cheng2017know}, Resnet-34, Resnet-18, Resnet-50 \citep{he2016deep}, ResNeSt-101 \citep{zhang2022resnest}, and MaxViT \citep{tu2022maxvit}, which are not explored in previous MI defense setups \citep{wang2021improving, peng2022bilateral}.

\begin{table}[ht]
\begin{adjustbox}{width=1.0\columnwidth,center}
\begin{tabular}{cccc}
\hline
\textbf{}     & \textbf{\#Classes} & \textbf{\#Images} & \textbf{\#Attack Classes} \\ \hline
CelebA        & 1,000              & 27,018            & 300                       \\
Facescrub     & 530                & 106,863           & 530                       \\
VGGFace2      & 8,631              & 3.31M             & 100                       \\
Stanford Dogs & 120                & 20,580            & 120                       \\ \hline
\end{tabular}
\end{adjustbox}

\caption{MI Private Dataset Setting. We follow previous works \cite{zhang2020secret,chen2021knowledge,wang2021variational,an2022mirror,nguyen_2023_CVPR,struppek2022plug} for the datasets selection.}
\label{tab:MI_Dataset_Setup}
% \vspace{-0.2cm}
\end{table}

\begin{table}[]
    \centering
    \begin{adjustbox}{width=1.0\columnwidth,center}
    \begin{tabular}{cccccccc}
    \hline
        \boldmath{$\lambda_x$} \T\B & 0.05 & 0.05 & 0.05 & 0.06 & 1.0 & 1.0 & 1.0 \\ \hline
        \boldmath{$\lambda_y$} \T\B & 0.5 & 0.4 & 0.6 & 0.5 & 0.5 & 5.0 & 10.0 \\ \hline
        \textbf{Natural Acc} & 80.35 \T\B & 73.69 & 76.46 & 76.13 & 23.27 & 57.57 & 57.04 \\ \hline
    \end{tabular}
    \end{adjustbox}
    \caption{The SOTA MI defense, BiDO is sensitive to hyper-parameters, posing challenges for applying effectively to different architectures of target classifier $T$ or private dataset $D_{priv}$. BiDO simultaneously optimizes two objectives: $d(x, f)$ (limiting information of input $x$ and feature representations $f$) and $d(f, y)$ (providing sufficient information about label $y$ to $f$), in addition to the main objectives $\mathcal{L}$. Therefore, the final objective is $\mathcal{L} + \lambda_x d(x, z) + \lambda_y d(f, y)$, where careful weight selection for $\lambda_x$ and $\lambda_y$ is necessary to achieve a balanced training among three objectives. It is clear that inappropriate values of $\lambda_x$ and $\lambda_y$ in BiDO cause an unstable training $T$. \textit{Note that \citep{peng2022bilateral} requires an extensive grid search to determine suitable values for $\lambda_x$ and $\lambda_y$}}
    \label{tab:SOTA_limitation}
    % \vspace{-1.0cm}

\end{table}

\textbf{Pre-trained Dataset for Target Classifier \boldmath{${\mathcal{D}_{pretrain}}$}.} We use Imagenet-1K \citep{deng2009imagenet} for VGG16, Resnet-18/50, ResNeSt-101, and MaxViT, and MS-CelebA-1M \citep{guo2016ms} for IR152 and FaceNe64, following previous works \citep{chen2021knowledge, zhang2020secret}. For Resnet-34, since it is trained from scratch in the original VMI setup \citep{wang2021variational}, we freeze the layers initialized from scratch. In Sec.~\ref{Pre-trained datasets}, we also study two additional pre-trained datasets, Facescrub \citep{ng2014data} and Pubfig83 \citep{pinto2011scaling}.

\textbf{MI Attack Method.} Our work focuses on white-box attacks, the most effective method in the literature. Following the SOTA MI defense \citep{peng2022bilateral}, we evaluate our proposed TL-DMI against three well-known attacks: GMI \citep{zhang2020secret}, KEDMI \citep{chen2021knowledge}, and VMI \citep{wang2021variational}. We further evaluate our proposed TL-DMI against current SOTA MI attacks as well as other SOTA MI attacks LOMMA \citep{nguyen_2023_CVPR}, PPA \citep{struppek2022plug}, and MIRROR \citep{an2022mirror}. The details for MI attack setups can be found below and in the Tab.~\ref{tab:MI_Attack_Setup}:
\begin{itemize} [leftmargin=0.5cm]
    \item \textbf{GMI} \citep{zhang2020secret} uses a pre-trained GAN to understand the image structure of an additional dataset. It then identifies inversion images by analyzing the latent vector of the generator.

    \item \textbf{KEDMI} \citep{chen2021knowledge}  expands on GMI \citep{zhang2020secret} by training a discriminator to differentiate between real and fake samples and predict the label as the target model. The authors also propose modeling the latent distribution to reduce inversion time and enhance the quality of reconstructed samples.

    \item \textbf{VMI} \citep{wang2021variational} introduces a probabilistic interpretation of MI and presents a variational objective to approximate the latent space of the target data.

    \item \textbf{LOMMA} \citep{nguyen_2023_CVPR} introduces two concepts of logit loss for identity loss and model augmentation to improve attack accuracy of previous MI attacks including GMI, KEDMI, and VMI.

    \item \textbf{PPA} \citep{struppek2022plug} proposes a framework for MI attack for high resolution images, which enable the use of a single GAN (i.e., StyleGAN) to attack a wide range of targets, requiring only minor adjustments to the attack. 

    \item \textbf{MIRROR} \citep{an2022mirror} proposes a MI attack framework based on StyleGAN similar to PPA, which aims at reconstructing private images having high fidelity. 

    \item \textbf{BREPMI} \citep{kahla2022label} introduce a new MI attack that can reconstruct private training data using only the predicted labels of the target model. The attack works by evaluating the predicted labels over a sphere and then estimating the direction to reach the centroid of the target class.
\end{itemize}

% \usepackage{multirow}
\begin{table}[ht]
\begin{adjustbox}{width=1.0\columnwidth,center}
\begin{tabular}{cccp{1.4cm}p{1.4cm}}
\hline
\multicolumn{1}{l}{} & \textbf{\#Iteration}  & \textbf{\boldmath{$w$} clipping} & \textbf{Learning rate} & \textbf{\#Attack per class} \T\B \\ \hline
\textbf{GMI}         & 3000                                 & Yes                 & 0.02                   & 5                              \T\B \\
\textbf{KEDMI}       & 3000                                  & Yes                 & 0.02                   & 5                              \T\B \\
\textbf{VMI}         & 320                                & -                   & 0.0001                 & 100                            \T\B \\

\textbf{LOMMA}         & 2400                               & Yes                   & 0.02                 & 5                            \T\B \\ 

\textbf{PPA}         & 50                    & Yes                   & 0.005                 & 50                            \T\B \\ 

\textbf{MIRROR}         & 500                        & Yes                   & 0.25                 & 8                            \T\B \\ 

\hline

\end{tabular}
\end{adjustbox}
\caption{MI Attack Setups. We follow the MI setups from previous works \cite{chen2021knowledge,peng2022bilateral,nguyen_2023_CVPR,struppek2022plug,an2022mirror}}
\label{tab:MI_Attack_Setup}
% \vspace{-0.5cm}
\end{table}

% \usepackage{multirow}
\begin{table}[ht]
\begin{adjustbox}{width=1.0\columnwidth,center}
\begin{tabular}{ccccccc}
\hline
\textbf{Architecture}      & \textbf{Method}      \T\B         & \boldmath{$\lambda_{MID}$} & \boldmath{$\lambda_{x}$} & \boldmath{$\lambda_{y}$} & \boldmath{$|\theta_C|$} & \textbf{Natual Acc} $\Uparrow$ \\ \hline
\multirow{12.5}{*}{VGG16}   
&  No. Def              & -                     & -                      & -                      & 16.8                     & 89.00           \T    \\
& MID              & 0.01                     & -                      & -                      & -                       & 68.39               \\
                           &    MID                               & 0.003                    & -                      & -                      & -                       & 78.70               \\
                           & BiDO-COCO        & -                        & 10                     & 50                     & -                       & 74.53               \\
                           &    BiDO-COCO                               & -                        & 5                      & 50                     & -                       & 81.55               \\
                           & BiDO-HSIC        & -                        & 0.05                   & 1                      & -                       & 70.31               \\
                           &      BiDO-HSIC                             & -                        & 0.05                   & 0.5                    & -                       & 80.35               \\
                           & TL-DMI             & -                        & -                      & -                      & 15.0                      & 86.57               \\
                           &        TL-DMI                           & -                        & -                      & -                      & 13.9                    & 83.41               \\
                           &        TL-DMI                           & -                        & -                      & -                      & 11.5                    & 77.89               \\
                           &        TL-DMI                           & -                        & -                      & -                      & 9.1                     & 69.80               \\
                           & TL-DMI + BiDO-HSIC & -                        & 0.05                   & 0.4                    & 15.0                      & 84.31               \\
                           &     TL-DMI + BiDO-HSIC                              & -                        & 0.03                   & 0.4                    & 15.0                      & 82.15               \B \\ \hline
\multirow{5.5}{*}{Resnet-34} 
& No. Def                               & -                        & -                      & -                      & 21.5                      & 69.27               \T \\
& MID                               & 0                        & -                      & -                      & -                       & 52.52                \\
                           & BiDO-COCO                         & -                        & 0.05                   & 2.5                    & -                       & 59.34               \\
                           & BIDO-HSIC                         & -                        & 0.1                    & 2                      & -                       & 61.14               \\
                           & TL-DMI                              & -                        & -                      & -                      & 21.1                    & 62.20               \B \\ \hline

\multirow{2.5}{*}{IR152}                  & No. Def                              & -                        & -                      & -                      & 62.6                    & 93.52               \T \\ \
                  & TL-DMI                              & -                        & -                      & -                      & 17.8                     & 86.70               \B \\ \hline

\multirow{2.5}{*}{FaceNet64}                  & No. Def                              & -                        & -                      & -                      & 35.4                    & 88.50               \T \\ \
                  & TL-DMI                              & -                        & -                      & -                      & 34.4                    & 83.61               \B \\ \hline

\multirow{2.5}{*}{Resnet-18}                    & No. Def                              & -                        & -                      & -                      & 11.7                    & 95.30               \T \\ 
                 & TL-DMI                              & -                        & -                      & -                      & 8.9                    & 91.17               \B \\ \hline

\multirow{2.5}{*}{MaxViT}                  & No. Def                              & -                        & -                      & -                      & 30.9                    & 96.57               \T \\ 
                  & TL-DMI                              & -                        & -                      & -                      & 18.3                    & 93.00               \B \\ \hline

\multirow{2.5}{*}{ResNeSt-101}                  & No. Def                              & -                        & -                      & -                      & 48.4                    & 75.07               \T \\ 
                 & TL-DMI                              & -                        & -                      & -                      & 27.9                    & 79.64               \B \\ \hline
\end{tabular}
\end{adjustbox}
% \vspace{-0.3cm}
\caption{Hyperparameters setting for training target classifiers. We follow previous work \cite{peng2022bilateral} for the hyperparameters selection of MID and BiDO.}
\label{tab:Hyperparam_Settings}
% \vspace{-0.6cm}
\end{table}

\subsection{Evaluation metrics}

% \usepackage{multirow}
\begin{table}[ht]
\begin{adjustbox}{width=\columnwidth,center}
% \usepackage{multirow}
\begin{tabular}{ccp{2.0cm}cc}
\hline
\textbf{Architecture}      & \textbf{Method}  & \textbf{Total Training Time (Seconds)} $\Downarrow$  & \textbf{Ratio} $\Downarrow$ & \textbf{Natural Acc} $\Uparrow$ \T\B \\ \hline
\multirow{5}{*}{VGG16}     & No. Def          & 2122                                                                             & 1.00           & 89.00                \T \\
                           & BiDO-COCO        & 3288                                                                             & 1.55           & 81.55                \\
                           & BiDO-HSIC        & 3296                                                                             & 1.55           & 80.35                \\
                           & \textbf{TL-DMI}             & \textbf{1460}                                                                             & \textbf{0.69}           & \textbf{83.41}                \\
                           & \textbf{TL-DMI + BiDO-HSIC} & \textbf{2032}                                                                             & \textbf{0.96}           & \textbf{84.14}                \B \\ \hline
\multirow{2}{*}{IR152}     & No. Def          & 6019                                                                             & 1.00           & 93.52                \T \\
                           & \textbf{TL-DMI}             & \textbf{2808}                                                                             & \textbf{0.47}           & \textbf{86.70}                \B \\ \hline
\multirow{2}{*}{FaceNet64} & No. Def          & 16344                                                                            & 1.00           & 88.50                \T \\
                           & \textbf{TL-DMI}             & \textbf{14448}                                                                            & \textbf{0.88}           & \textbf{83.61}                \B \\ \hline
\end{tabular}
\end{adjustbox}
\caption{Computational Resource. We remark that our proposed TL-DMI achieve SOTA MI robustness while reduce the computational cost as we keep the same training protocol and update fewer parameters than No. Def and SOTA MI Defense BiDO.}
\label{tab:Computation_Resource}
% \vspace{-0.5cm}
\end{table}

In the main manuscript, we make use of Natural Accuracy, Attack Accuracy, and K-Nearest-Neighbors Distance (KNN Dist) metrics to evaluate MI robustness. These metrics are described as:

\begin{itemize} [leftmargin=0.5cm]
      \item \textbf{Attack accuracy (AttAcc).} To gauge the effectiveness of an attack, we develop an \textit{evaluation classifier} that predicts the identities of the reconstructed images. This metric assesses the similarity between the generated samples and the target class. If the evaluation classifier attains high accuracy, the attack is considered successful. To ensure an unbiased and informative evaluation, the evaluation classifier should exhibit maximal accuracy.

      \item \textbf{Natural accuracy (Acc).} In addition to assessing the Attack Acc of a released model, it is also necessary to ensure that the model performs satisfactorily in terms of its classification utility. The evaluation of the model's classification utility is typically measured by its natural accuracy, which refers to the accuracy of the model in the classification problem.
    
      \item \textbf{K-Nearest Neighbors Distance (KNN Dist).} The KNN Dist metric provides information about the proximity between a reconstructed image associated with a particular label or ID, and the images that exist in the private training dataset. This metric is calculated by determining the shortest feature distance between the reconstructed image and the actual images in the private dataset that correspond to the given class or ID. To calculate the KNN Dist, an $l_2$ distance measure is used between the two images in the feature space, specifically in the penultimate layer of the evaluation model. This distance measure provides insight into the similarity between the reconstructed and the real images in the training dataset for a particular label or ID.

      \item \textbf{\boldmath{$\delta_{EvalNet}$} and \boldmath{$\delta_{FaceNet}$}} These metrics are measured by the squared $l_2$ distance between the activation in the penultimate layers. $\delta_{EvalNet}$ is computed via Evaluation Model while $\delta_{EvalNet}$ is computed via pre-trained FaceNet \citep{schroff2015facenet}. A lower value indicates that the attack results are more visually similar to the training data.

      \item \textbf{\boldmath{$\ell_2$} distance}. $\ell_2$ distance measures how similar the inverted images are to the private data by computing the distance between reconstructed features the centroid features of the private data. A lower distance means that the inverted images are more similar to the target class.

      \item \textbf{Frechet inception distance (FID)}. FID is commonly used to evaluate generative model to access the generated images. The FID measures the similarity between two sets of images by computing the distance between their feature vectors. Feature vectors are extracted using an Inception-v3 model that has been trained on the ImageNet dataset. In the context of MI, a lower FID score indicates that the reconstructed images are more similar to the private training images.
\end{itemize}

\section{Reproducibility} \label{Reproducibility}

\begin{figure*}[ht]
\centering
\begin{adjustbox}{width=1.0\textwidth,center}
\includegraphics[width=1.0\textwidth]{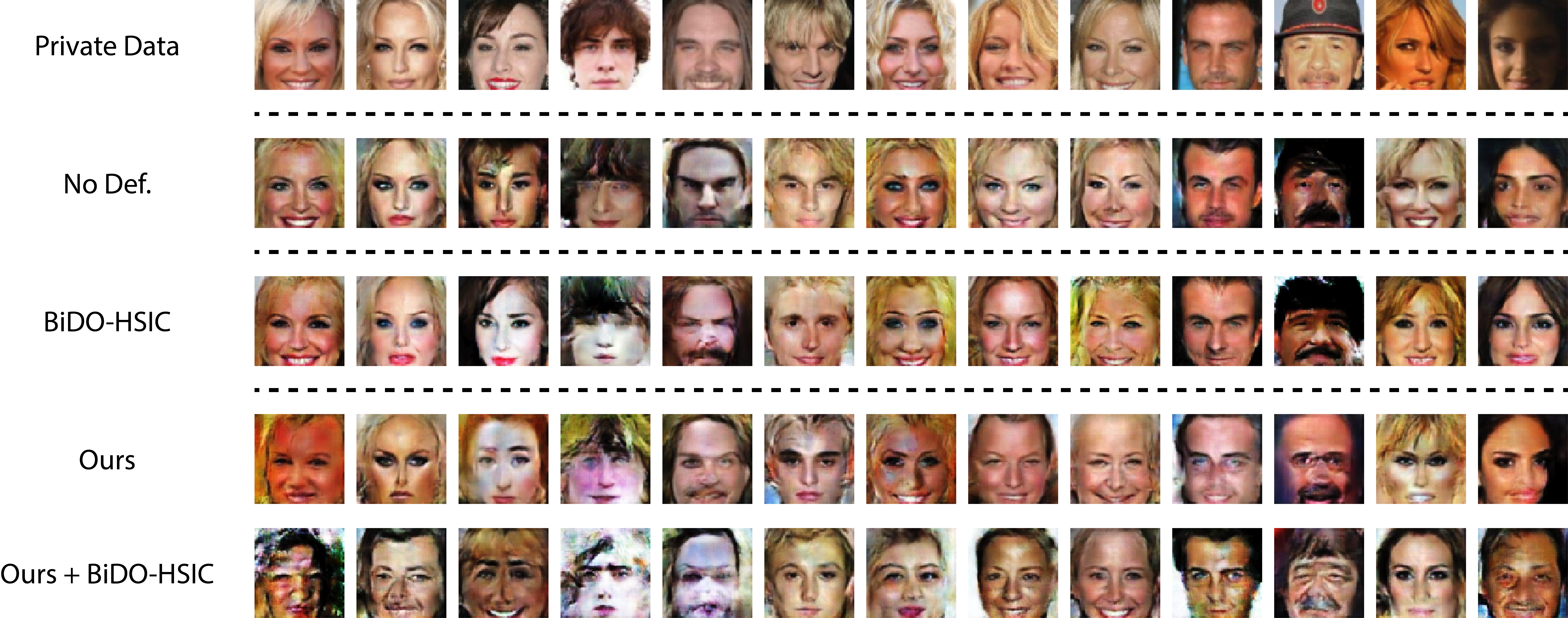}
\end{adjustbox}
\caption{Qualitative results to showcase the effectiveness of our proposed TL-DMI, using KEDMI \citep{chen2021knowledge} with $\mathcal{D}_{priv}$ = CelebA, $\mathcal{D}_{pub}$ = CelebA, $\mathcal{D}_{pretrain}$ = Imagenet1K, and $T$ = VGG16. The visual comparison reveals that our proposed TL-DMI achieves competitive reconstruction of private data, while the hybrid approach combining our method with BiDO-HSIC demonstrates a significant degradation in MI attack and reconstruction quality.}
% \vspace{-0.4cm}
\label{fig:Visualization}
\end{figure*}

\subsection{The details for training $T$}

\textbf{Training target classifier \boldmath{$T$}.} In this work, we employ VGG16 \citep{simonyan2014very}, IR152 \citep{he2016deep}, and FaceNet64  \citep{cheng2017know} for our investigation. All target classifiers are trained on CelebA dataset. For GMI \citep{zhang2020secret} and KEDMI \citep{chen2021knowledge}, the target classifiers trained were VGG16, IR152, and FaceNet64, while Resnet-34 was used as the target classifier for VMI \citep{wang2021variational}. As mentioned in the main manuscript, we employ Imagenet-1K as the pre-trained dataset for VGG16, while MS-CelebA-1M was used as the pre-trained dataset for IR152 and FaceNet64. The details of the training procedure are shown in Tab.~\ref{tab:training_settings} below.

\begin{figure}[t]
    % \centering
    % \renewcommand\thefigure{A}
    \begin{tabular}{c}
    % \vspace{-0.2cm}
    \includegraphics[width=0.47\textwidth]{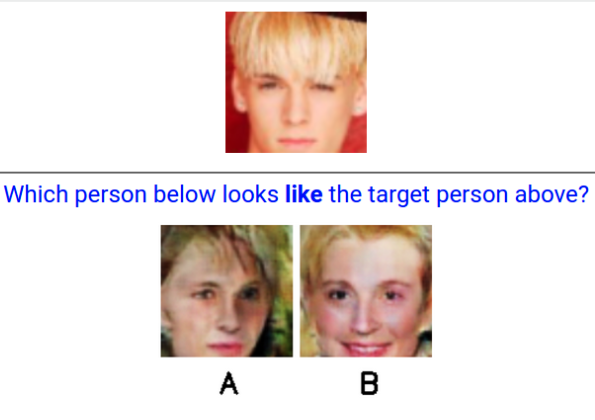} \\
    % \caption{Interface}
    % \midrule
    \begin{adjustbox}{width=0.47\textwidth,center}
        \begin{tabular}{c|c|c|c}
        % \multicolumn{4}{c}
        % {\textbf{(ii) $D_{priv}$ = CelebA / $D_{pub}$ = CelebA / $T$ = FaceNet64}
        % } \\ 
        \toprule
        Method & NaturalAcc $\Uparrow$ & AttackAcc $\Downarrow$ & User Preference $\Downarrow$ \\ \midrule
        BiDO & 80.35\%  & 45.80\% & 87.1\% \\ \midrule
        \bf{BiDO+TL-DMI} &\bf{82.15\%} &\bf{17.40\%} &\bf{12.9\%}\\
        \bottomrule
        \end{tabular}
        
    \end{adjustbox}
    
    \end{tabular}
    
 % \vspace{-0.5cm}
  \caption{
  % {User study inference and result}
  {An example for the user study inference (top) and the results for user study (bottom). Compared to BiDO, our proposed TL-DMI provide a better defense with higher natural accuracy but lower user preference.}
  }
\label{fig:penultimate-layer-visualization}
% \vspace{-0.5cm}
\end{figure}

\textbf{Important Hyper-parameters.}
In our work, we performed an analysis of our proposed TL-DMI against existing SOTA model inversion defense methods: MID \citep{wang2021improving} and Bilateral Dependency Optimization (BiDO)\citep{peng2022bilateral}. MID \citep{wang2021improving} adds a regularizer $d(x, T(x))$ to the main objective during the target classifier’s training to penalize the mutual information between inputs $x$ and outputs $T(x)$. BiDO \citep{peng2022bilateral} attempts to minimize $d(x, z)$ to reduce the amount of information about inputs $x$ embedded in feature representations $z$, while maximizing $d(z, y)$ to provide $z$ with enough information about $y$ to restore the natural accuracy. 
For simplicity, we use $\lambda_{MID}$, $\lambda_x$, and $\lambda_y$ to represent $d(x, T(x))$, $d(x, z)$, and $d(z, y)$ respectively. The settings of these hyper-parameters are detailed in Tab.~\ref{tab:Hyperparam_Settings}.

\subsection{Compute resource} \label{Computing resource}

All our experiments are run on NVIDIA RTX A5000 GPUs. Given that our work is focused on model inversion defense, we provide the total training time (seconds) for the target classifier and the ratio of training time between each model inversion defense method against the No. Def. The results in Tab.~\ref{tab:Computation_Resource} below show that \textbf{our proposed TL-DMI can greatly reduce the amount of time required to train the target classifier.}

\subsection{Error Bars} \label{error bar}

For this section, we ran a total of 7 setups (3 times for each setup) across 4 different architectures of the target classifiers, and report their respective natural accuracy and attack accuracy values. For each experiment, we use the same MI attack setup and training settings for target classifiers as reported in the main setups comparing with BiDO and Tab.~\ref{tab:training_settings} respectively. We show that the results obtained are reproducible and do not deviate much from the reported values in the main paper. These results can be found in Tab.~\ref{tab:error_bar} below.

% \vspace{-0.3cm}
\section{Qualitative results} \label{Qualitative result}

\subsection{Visual Comparison} We evaluate the efficacy of our proposed TL-DMI along with BiDO for preventing privacy leakage on CelebA and also provide visualisation of the samples produced using the KEDMI \citep{chen2021knowledge} MI attack method. In Fig.~\ref{fig:Visualization} below, each column represents the same identity and the first row represents the ground-truth private data while each subsequent row shows the attack samples reconstructed for each MI defense method.

\subsection{User study} We conduct our user study via Amazon MTurk with the interface as shown above. We adapt our user study from MIRROR. In the setup, participants are presented with a real image of the target class, and then asked to pick one of two inverted images that is more closely aligned with the real image. The order is randomized, with each image pair displayed on-screen for a maximum duration of 60 seconds. The assessment encompassed all 300 targeted classes. Each pair of inverted images is assigned to 10 unique individuals, thus our user study involves a total of 3000 pairs of inverted images. We use KEDMI as the MI attack with $\mathcal{D}_{priv}=CelebA$, $\mathcal{D}_{pub}=CelebA$, $T=FaceNet~$. \textit{Consistent with the AttackAcc, the user study shows that our proposed TL-DMI provides better defense against the reconstruction of private data characteristics compared to BIDO.}

\end{document}